\documentclass[preprint,12pt]{elsarticle}

\usepackage{array}
\newcolumntype{P}[1]{>{\centering\arraybackslash}p{#1}}
\newcolumntype{M}[1]{>{\centering\arraybackslash}m{#1}}
\usepackage{multirow}

\usepackage[utf8x]{inputenc}

\usepackage{array}
\usepackage{booktabs}
\newlength{\ColWidthNormal} 
\newlength{\ColWidthRowHeader} 

\newlength{\RuleOffsetLeft} 
\newlength{\RuleThicknessNormal} 

\newcolumntype{C}{>{\centering\arraybackslash\leavevmode}p{\ColWidthNormal}}
\newcolumntype{R}{>{\raggedleft\arraybackslash}p{\ColWidthNormal}}
\newcolumntype{L}{>{\raggedright\arraybackslash}p{\ColWidthRowHeader}}

\usepackage[dvipsnames]{xcolor}

\usepackage{amsmath}

\usepackage[caption=false]{subfig}

\usepackage{multirow}  % Used for multi-row merging in tables
\usepackage{array}        % Used for `m{}` command in tables

\usepackage{adjustbox}
\usepackage[graphicx]{realboxes}
\usepackage{rotating}
\usepackage{hyperref}
\usepackage{amssymb}
\journal{Information Fusion}

\begin{document}

\begin{frontmatter}



\title{Machine Learning Meets Advanced Robotic Manipulation}


\author[label1,label2]{Saeid Nahavandi}
\ead{snahavandi@swin.edu.au}
\author[label3]{Roohallah Alizadehsani\corref{cor1}%\fnref{label6}
}
\ead{r.alizadehsani@deakin.edu.au}
%\fntext[label6]{aaaa}
\cortext[cor1]{Corresponding author}
\author[label3]{Darius Nahavandi}
\ead{darius.nahavandi@deakin.edu.au}
\author[label3]{Chee Peng Lim}
\ead{chee.lim@deakin.edu.au}
\author[label4]{Kevin Kelly}
\ead{KEKELLY@tcd.ie}
\author[label5]{Fernando Bello}
\ead{F.Bello@imperial.ac.uk}

\affiliation[label1]{organization={Distinguished Professor, Associate Deputy Vice-Chancellor Research, Swinburne University of Technology},%Department and Organization
            %addressline={}, 
            city={Hawthorn},
            %postcode={}, 
            state={VIC 3122},
            country={Australia}}
            
\affiliation[label2]{organization={Harvard Paulson School of Engineering and Applied Sciences, Harvard University},%Department and Organization
            %addressline={}, 
            city={Allston},
            %postcode={}, 
            state={MA 02134},
            country=USA{}}

\affiliation[label3]{organization={Institute for Intelligent Systems Research and Innovation (IISRI), Deakin University},
             %addressline={},
             %city={},
             %postcode={},
             %state={},
             country={Australia}}
             
\affiliation[label4]{organization={Department of Mechanical \& Manufacturing Engineering, Trinity College},
             %addressline={},
             city={Dublin},
             %postcode={},
             %state={},
             country={Ireland}}
             
\affiliation[label5]{organization={Center for Engagement and Simulation Science, Department of Surgery and Cancer, Imperial College London},
%addressline={},
city={London},
%postcode={},
%state={},
country={United Kingdom}
}

\begin{abstract}
Automated industries lead to high quality production, lower manufacturing cost and better utilization of human resources. Robotic manipulator arms have major role in the automation process. However, for complex manipulation tasks, hard coding efficient and safe trajectories is challenging and time consuming. Machine learning methods have the potential to learn such controllers based on expert demonstrations. Despite promising advances, better approaches must be developed to improve safety, reliability, and efficiency of ML methods in both training and deployment phases. This survey aims to review cutting edge technologies and recent trends on ML methods applied to real-world manipulation tasks. After reviewing the related background on ML, the rest of the paper is devoted to ML applications in different domains such as industry, healthcare, agriculture, space, military, and search and rescue. The paper is closed with important research directions for future works.
\end{abstract}




\end{frontmatter}


\section{Introduction}
Manipulating different objects is a trivial task for humans owing to having unique vision capabilities and a highly flexible body structure with 224 degrees of freedom (DoF) and approximately 630 skeletal muscles \cite{zatsiorsky2012biomechanics}. However, robotic manipulation is one of the complex problems, and is an active research field. The reason is that replicating the vision and actuation capabilities of natural organisms (e.g. humans) is a very challenging task by using the respective (possibly simplified) robotic counterparts. As complexity of robot structures increases from hard to soft (section \ref{manip-struct-sec}), developing controllers for them becomes more difficult which is the motivation behind using machine learning (ML) methods for robot control development.

The robotic literature contains multiple surveys to track the rapid progress of ML methods specially reinforcement learning (RL) and deep learning (DL) in robotic problems. Amarjyoti \cite{amarjyoti2017deep} focused on the RL and deep RL (DRL) methods applied to learning robotic manipulation without investigating field-specific applications. Moreover, Amarjyoti \cite{amarjyoti2017deep} only considered discrete robots and neglected soft ones. Thuruthel et al. \cite{george2018control} and Kim et al. \cite{kim2021review} covered this shortcoming by investigating control strategies for soft manipulators.

Given the complexity of robotic manipulation, learning controllers from scratch is challenging and sometimes costly. Moreover, adapting to sudden changes and performing multi-tasking is not trivial. Imitation learning can be used to reproduce expert's behaviors on the robot and generalize to new environments \cite{fang2019survey}. Demonstrations for imitation learning are either direct or indirect. Training data collection in direct demonstration like kinesthetic teaching \cite{ragaglia2018robot} and tele-operation teaching \cite{zhang2018deep} is accomplished on the robot itself. Indirect demonstrations are collected in a separate environment without interaction with the robot \cite{wan2017teaching}. Another approach to ease the learning process is training in simulated environments and transferring the learned skills to real robots. This can be achieved by using simulation-to-real methods which have been reviewed extensively \cite{karoly2020deep, zhao2020sim, han2023survey}.

One of the stepping stones toward robust robotic manipulation is environment perception. Clearly, object manipulation is not feasible if the object's position cannot be determined with respect to the robot end-effector (EE). This requirement motivated Kleeberger et al. \cite{kleeberger2020survey} to review ML for object grasping methods based on vision. Vision-based robot grasping methods are either analytic (also denoted as geometric) or data-driven \cite{kleeberger2020survey, sahbani2012overview}. In analytic methods, force-closure grasp is performed with a multi-fingered robotic hand such that the grasp capability has four properties namely dexterity, equilibrium, stability, and certain dynamic behavior \cite{shimoga1996robot}. A grasp is called force-closure if and only if arbitrary force and moment can be exerted on the object using the robot finger tips without breaking contact with the object \cite{1087483}. To perform such a grasp, a constrained optimization problem is solved to satisfy a subset or all of these four properties. The second category of grasping methods is data-driven. This category is based on ranking sampled grasp candidates according to a specific measure. Moreover, the candidate generation is usually based on a heuristic grasp experience or one generated using a simulated or real robot and grasp skill is learned using ML. The focus of Kleeberger et al. \cite{kleeberger2020survey} is on data-driven methods. Newbury et al. \cite{newbury2023deep} also reviewed the two grasping categories with a special focus on DL solutions.

While developing state-of-the-art ML methods for learning complex problems is worthwhile, trying to use them in practice is just as important. That is why instead of focusing on ML methods themselves, Fabisch et al. \cite{fabisch2019survey} devoted a complete survey to real-world applications of ML methods for robot behavior learning. Following a similar approach, Benotsmane et al. \cite{benotsmane2020survey} investigated ML application to the industrial domain whereas \cite{saleem2021automation} investigated ML and DL methods for agricultural automation. These methods have even been used in space applications \cite{dai2022review}.

Given the versatility of learning problems in manipulation tasks, Kroemer et al. \cite{kroemer2021review} presented a formalization with five categories that covers majority of manipulation learning problems for soft and discrete robots. The five categories are state space representation learning, dynamics model learning, motor skills learning, preconditions learning for already learned skills, and hierarchical learning. The preconditions of a skill refer to the circumstances under which that skill can be executed. In hierarchical RL literature, the preconditions are also known as the initiation set of temporally extended actions which are called options \cite{sutton1999between}.

While the surveys listed above are useful in their own rights, we believe a new survey covering practical application of ML and DL to different domains is still needed. In this paper, we strive to review cutting edge ML/DL technologies in medical, industrial, agricultural, search and rescue, military, and space applications. Table \ref{compare-tbl} illustrates the contribution of our paper, as compared with existing survey papers.

The rest of the paper is as follows. Sections \ref{manip-struct-sec} and \ref{manip_control_sec} are devoted to robotic manipulator structures and control, respectively. RL basics are reviewed in section \ref{rl-primer-sec} and their usage and challenges in robotic manipulation are reviewed in section \ref{control-using-rl}. Similarly, applications of DL to manipulators are presented in section \ref{dl-manip-sec}. Sim-to-real methods related to manipulators are reviewed in section \ref{sim2real-sec}. Applications of manipulators to various domains are reviewed in section \ref{manip-in-action-sec}. Concluding remarks and suggestions for further research are presented in sections \ref{future-work-sec} and \ref{conc-sec}, respectively.

\begin{table}[]
\centering
\caption{Comparison with previous surveys on robotic manipulation: MDC, IDL, AGL, SRS, MLT, and SPC stand for medical, industrial, agricultural, search and rescue, military, and space, respectively.}
\begin{adjustbox}{width=1\textwidth}
\label{compare-tbl}
\begin{tabular}{|p{0.05\textwidth}|cccccc|p{0.05\textwidth}|p{0.35\textwidth}|}
\hline
\multicolumn{1}{|c|}{\multirow{2}{*}{
\begin{tabular}{@{}c@{}}
Ref,\\year
\end{tabular}
}} & \multicolumn{6}{c|}{Applications}                                                                                                                                                                       & \multicolumn{1}{c|}{\multirow{2}{*}{Robot types}} & \multicolumn{1}{c|}{\multirow{2}{*}{Reviewed methods}} \\ \cline{2-7}
\multicolumn{1}{|c|}{}               & \multicolumn{1}{c|}{MDC} & \multicolumn{1}{c|}{IDL} & \multicolumn{1}{c|}{AGL} & \multicolumn{1}{c|}{SRS} & \multicolumn{1}{c|}{MLT} & \multicolumn{1}{c|}{SPC} & \multicolumn{1}{c|}{}                             & \multicolumn{1}{c|}{}                                  \\ \hline
\cite{amarjyoti2017deep}, 2017 &
 \multicolumn{1}{l|}{-}      &
  \multicolumn{1}{l|}{-}     &
   \multicolumn{1}{l|}{-}    &
    \multicolumn{1}{l|}{-}   &
     \multicolumn{1}{l|}{-}  &
      -                      &
       discrete               & RL, DRL      \\ \hline
\cite{george2018control}, 2018                                       & \multicolumn{1}{l|}{\checkmark}     & \multicolumn{1}{l|}{\checkmark}        & \multicolumn{1}{l|}{-}           & \multicolumn{1}{l|}{-}               & \multicolumn{1}{l|}{\checkmark}      & -                      & soft                                              & ML, RL, fuzzy logic                        \\ \hline
\cite{fang2019survey}, 2019                                       & \multicolumn{1}{l|}{-}        & \multicolumn{1}{l|}{-}           & \multicolumn{1}{l|}{-}             & \multicolumn{1}{l|}{-}                 & \multicolumn{1}{l|}{-}   &     -     &   discrete   &  Imitation learning (behavioral cloning, inverse RL, adversarial imitation learning)          \\ \hline
\cite{fabisch2019survey}, 2019                                       & \multicolumn{1}{l|}{\checkmark}        & \multicolumn{1}{l|}{\checkmark}           & \multicolumn{1}{l|}{-}             & \multicolumn{1}{l|}{\checkmark}                 & \multicolumn{1}{l|}{-}         &      -                      &        discrete                                           &   RL, HRL, DRL                                                     \\ \hline
\cite{benotsmane2020survey}, 2020                                       & \multicolumn{1}{l|}{-}    & \multicolumn{1}{l|}{\checkmark}           & \multicolumn{1}{l|}{-}             & \multicolumn{1}{l|}{-}                 & \multicolumn{1}{l|}{-}         &              -              &              discrete           & RL, SL, EA, Fuzzy logic, search algorithms \\ \hline
\cite{kleeberger2020survey}, 2020                                       & \multicolumn{1}{l|}{-}        & \multicolumn{1}{l|}{\checkmark}           & \multicolumn{1}{l|}{-}             & \multicolumn{1}{l|}{-}                 & \multicolumn{1}{l|}{-}         &        -                    &  discrete                                                 &         RL, SL, DL, sim2real                                               \\ \hline
\cite{kroemer2021review}, 2021                                       & \multicolumn{1}{l|}{-}      & \multicolumn{1}{l|}{-}         & \multicolumn{1}{l|}{-}           & \multicolumn{1}{l|}{-}               & \multicolumn{1}{l|}{-}       &             -               & discrete, soft & RL, HRL, DL, imitation learning, segmentation, supervised/unsupervised learning
                                           \\ \hline
\cite{saleem2021automation}, 2021    & \multicolumn{1}{l|}{-}        & \multicolumn{1}{l|}{-}           & \multicolumn{1}{l|}{\checkmark}             & \multicolumn{1}{l|}{-}                 & \multicolumn{1}{l|}{-}         &      -                    &       discrete          &    DL, ML(KNN, SVM, LDA, …)\\ \hline
\cite{liu2021deep}, 2021    & \multicolumn{1}{l|}{-}        & \multicolumn{1}{l|}{-}           & \multicolumn{1}{l|}{-}             & \multicolumn{1}{l|}{-}                 & \multicolumn{1}{l|}{-}         &   -                     &         discrete         &      DRL, imitation learning, meta learning                    \\ \hline
\cite{kim2021review}, 2021                                       & \multicolumn{1}{l|}{\checkmark}        & \multicolumn{1}{l|}{-}           & \multicolumn{1}{l|}{-}             & \multicolumn{1}{l|}{-}                 & \multicolumn{1}{l|}{-}         &      -         &   soft            &                     ML(SVM, KNN, …), DL, RL                                 \\ \hline
\cite{karoly2020deep}, 2021  & \multicolumn{1}{l|}{-}        & \multicolumn{1}{l|}{-}           & \multicolumn{1}{l|}{-}             & \multicolumn{1}{l|}{-}                 & \multicolumn{1}{l|}{-}         &     -    & discrete & DL, RL, sim2real, unsupervised learning, semi-supervised learning, transfer learning\\ \hline
\cite{zhao2020sim}, 2021   & \multicolumn{1}{l|}{-} & \multicolumn{1}{l|}{-} & \multicolumn{1}{l|}{-}   & \multicolumn{1}{l|}{\checkmark}             & \multicolumn{1}{l|}{-}         &      -    & discrete   & RL, DRL, sim2real, imitation learning, dynamics randomization          \\ \hline
\cite{dai2022review}, 2022  & \multicolumn{1}{l|}{-}      & \multicolumn{1}{l|}{-}         & \multicolumn{1}{l|}{-}           & \multicolumn{1}{l|}{-}               & \multicolumn{1}{l|}{-}       & \checkmark                        & discrete                                          & RL, DL, interpolation, evolutionary methods            \\ \hline
\cite{newbury2023deep}, 2023  & \multicolumn{1}{l|}{-}        & \multicolumn{1}{l|}{\checkmark}  & \multicolumn{1}{l|}{\checkmark}  & \multicolumn{1}{l|}{-}   & \multicolumn{1}{l|}{-}  & - & discrete, soft     &    RL, DL, ML(GMM, SVM, GP)     \\ \hline
\cite{han2023survey}, 2023                                       & \multicolumn{1}{l|}{-}      & \multicolumn{1}{l|}{\checkmark}        & \multicolumn{1}{l|}{-}           & \multicolumn{1}{l|}{-}               & \multicolumn{1}{l|}{-}       & -                         & discrete                                          & RL, sim2real, imitation/transfer learning              \\ \hline
ours                                                                                & \multicolumn{1}{l|}{\checkmark}     & \multicolumn{1}{l|}{\checkmark}        & \multicolumn{1}{l|}{\checkmark}          & \multicolumn{1}{l|}{\checkmark}              & \multicolumn{1}{l|}{\checkmark}      & \checkmark   & discrete, soft               & RL, HRL, DL, sim2real, dynamics randomization \\ \hline
\end{tabular}
\end{adjustbox}
\end{table}

\section{Manipulators structure}\label{manip-struct-sec}
\begin{figure}[] % use figure* instead of figure if you want to span subfigures across two columns
\centering
\subfloat[]{\includegraphics[width=0.13\textwidth]{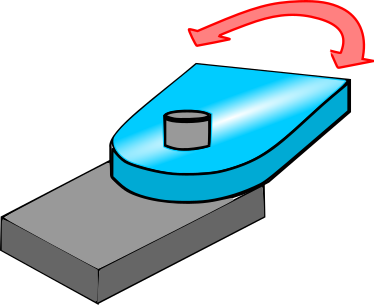}%2.6
\label{revolute-joint-fig}}
\hspace*{0.9em} % set the horizontal space between side by side figures
\subfloat[]{\includegraphics[width=0.13\textwidth]{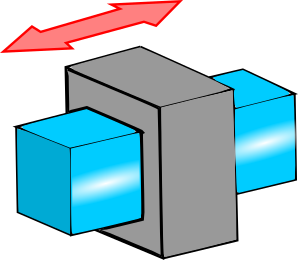}%5.2
\label{prismatic-joint-fig}}
\hspace*{0.9em}
\subfloat[]{\includegraphics[width=0.13\textwidth]{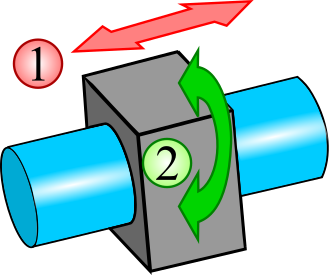}%
\label{cylindrical-joint-fig}}
\subfloat[]{\includegraphics[width=0.16\textwidth]{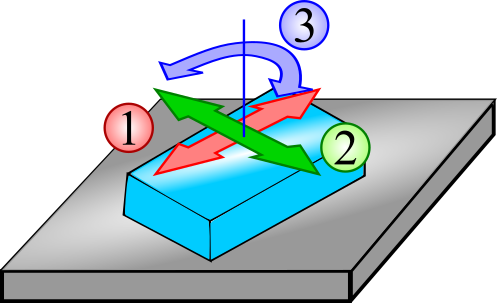}%
\label{planar-joint-fig}}
\hspace*{0.9em}
\subfloat[]{\includegraphics[width=0.13\textwidth]{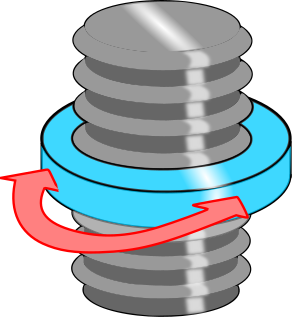}%
\label{screw-joint-fig}}
\hspace*{0.9em}
\subfloat[]{\includegraphics[width=0.16\textwidth]{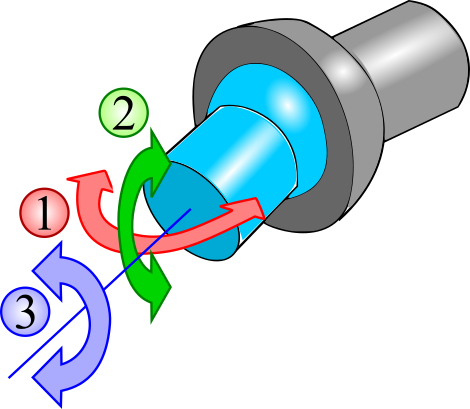}%
\label{spherical-joint-fig}}
\caption{Different types of joints: (a) revolute joint with 1 DoF, (b) prismatic joint with 1 DoF, (c) cylindrical joint with 2 DoF, (d) planar joint with 3 DoF, (e) screw joint with 1 DoF, (f) spherical joint with 3 DoF.}
\label{joint-fig}			% label for the whole figure
\end{figure}

Traditionally, manipulators consist of multiple links connected together with multiple joints. The number and configuration of links and joints depend on the tasks for which manipulators are designed for. Different types of joints are shown in Fig. \ref{joint-fig} \cite{craig2006introduction}. Revolute joint (Fig. \ref{revolute-joint-fig}), prismatic joint (Fig. \ref{prismatic-joint-fig}), and screw joint (Fig. \ref{screw-joint-fig}) provide rotational, translational, and rotational motion around one axis, respectively. These three joints have only one DoF. Cylindrical joint (Fig. \ref{cylindrical-joint-fig}) has translation and rotational motions providing two DoF. Planar joint (Fig. \ref{planar-joint-fig}) provides translation along two axes and rotation around one axis yielding three DoF. Spherical joint (Fig. \ref{spherical-joint-fig}) provides rotation motions around three axes, producing three DoF.  % TODO: talk about joints

\begin{figure}[] % use figure* instead of figure if you want to span subfigures across two columns
\centering
\subfloat[]{\includegraphics[width=0.16\textwidth]{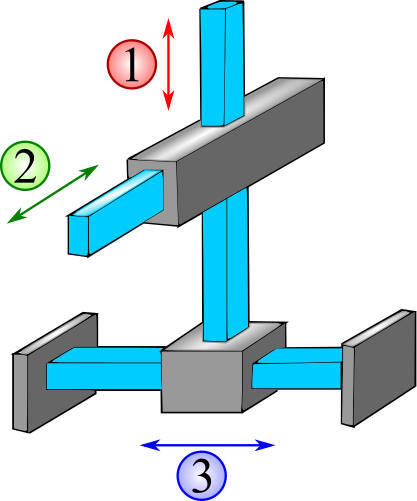}%
\label{cartesian-robot-fig}}
%\hspace*{0.9em} % set the horizontal space between side by side figures
\subfloat[]{\includegraphics[width=0.17\textwidth]{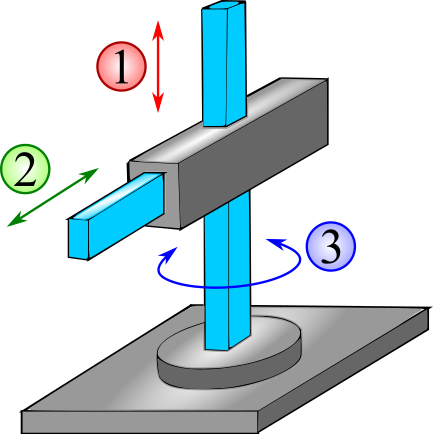}%
\label{cylindrical-robot-fig}}
%\hspace*{0.9em}
\subfloat[]{\includegraphics[width=0.185\textwidth]{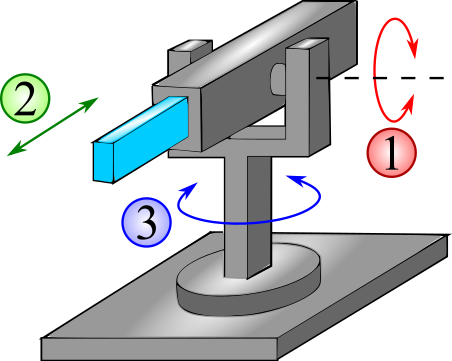}%
\label{spheric-robot-fig}}
\subfloat[]{\includegraphics[width=0.23\textwidth]{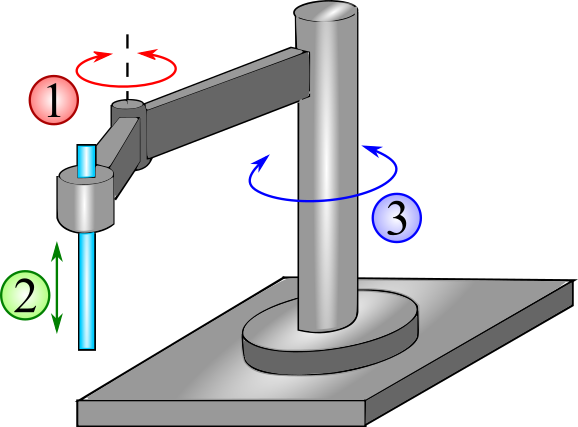}%
\label{scara-robot-fig}}
%\hspace*{0.9em}
\subfloat[]{\includegraphics[width=0.19\textwidth]{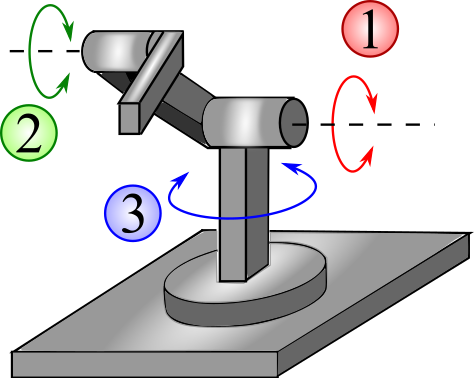}%
\label{angular-robot-fig}}
\caption{Examples of robot arm configurations using different joint types: (a) cartesian robot, (b) cylindrical robot, (c) spherical robot, (d) SCARA robot, and (e) angular robot.}
\label{different-robot-arm-configs-fig}			% label for the whole figure
\end{figure}

Utilizing different combinations of joints illustrated in Fig. \ref{joint-fig} leads to different manipulator arms capable of handling various tasks. Some typical manipulator arms using different joint types are shown in Fig. \ref{different-robot-arm-configs-fig} \cite{rosales2002forward}. Cartesian robot (Fig. \ref{cartesian-robot-fig}) is only made of prismatic joints providing translational motion along three axes. Cylindrical robot (Fig. \ref{cylindrical-robot-fig}) makes use of a cylindrical joint and a prismatic one to provide translation along two axes and rotation around one axis. Spherical robot (Fig. \ref{spheric-robot-fig}) is able to cover a half-sphere due to its two revolute joints and one prismatic one. SCARA \cite{al2012scara} (Fig. \ref{scara-robot-fig}) stands for Selective Compliance Assembly Robot Arm which is equipped with two revolute joints and a prismatic one. The operational space of SCARA is similar to a cylinder. Angular robot (Fig. \ref{angular-robot-fig}) utilizes three revolute joints to provide a half-sphere operational space.
% TODO: talk about different manip arms based on using different combination of joints

The robotic arms made by chaining multiple links together through joints are called hard robots\cite{george2018control}. Maneuverability of hard robots is limited by the number, type, and configuration of joints used in them. Using specific joints configuration eases computation of the manipulator end-effector pose (forward kinematics) and manipulator motion planning (inverse kinematics). However, hard robots with precise movements \cite{kuka-iiwa, elle1995mechanical} require high-end servo motors, which are costly. Additionally, in human-computer interaction scenarios, the operational space of a hard robot must be specified very carefully to avoid collision of the robot with its user during operation. These issues of hard robots have pushed researchers toward the development of soft robots. As mentioned by Chen et al. \cite{chen2022model}, soft robots are low cost, easily manufactured, flexible, and compliant with human safety \cite{walker2020soft, shintake2018soft, zhang2020state, sinatra2019ultragentle}. 

\begin{figure}[] 
\centering
\includegraphics[width=1\textwidth]{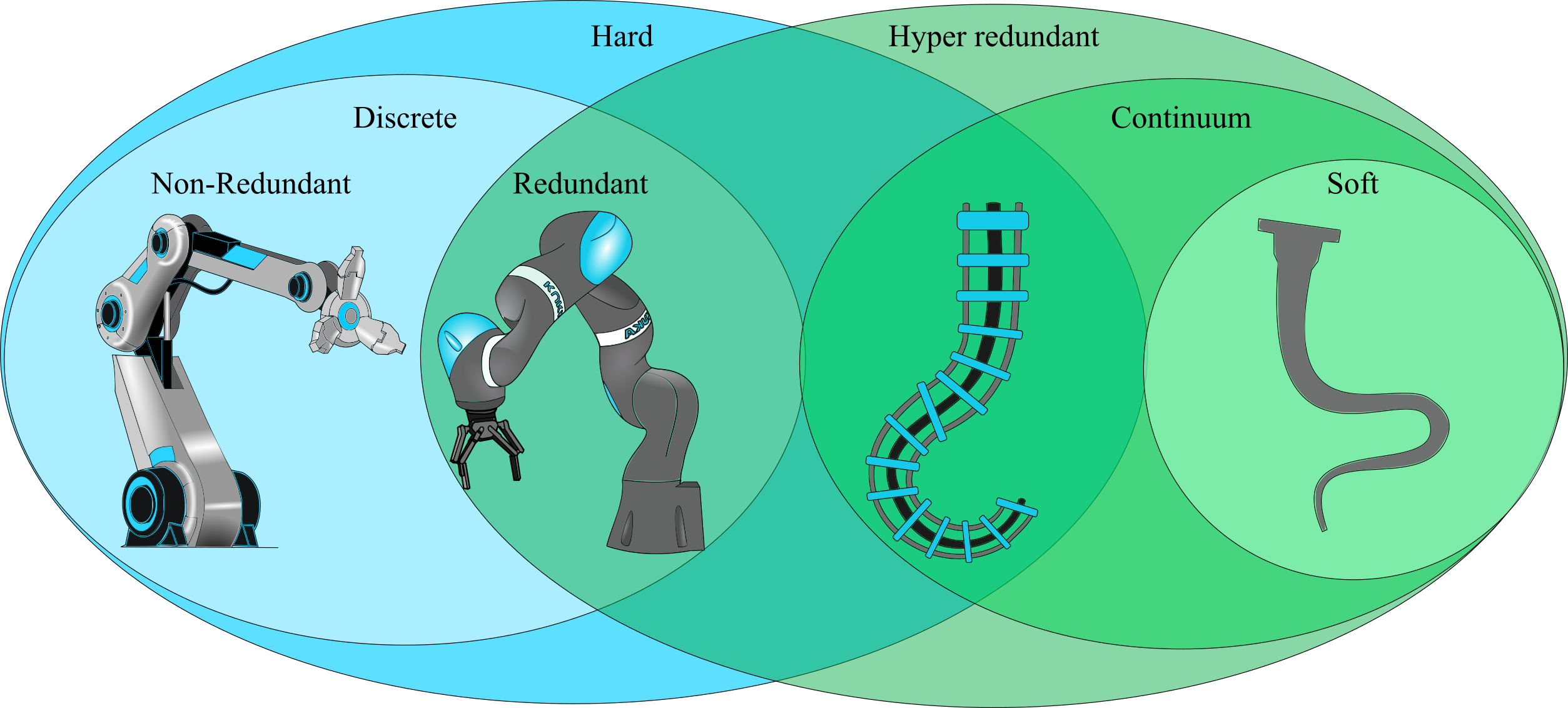}
\caption{Classification of manipulators from hard to soft.}
\label{hard2soft-robot-range-fig}			% label for the whole figure
\end{figure}

\begin{figure}[] 
\centering
\includegraphics[width=0.5\textwidth]{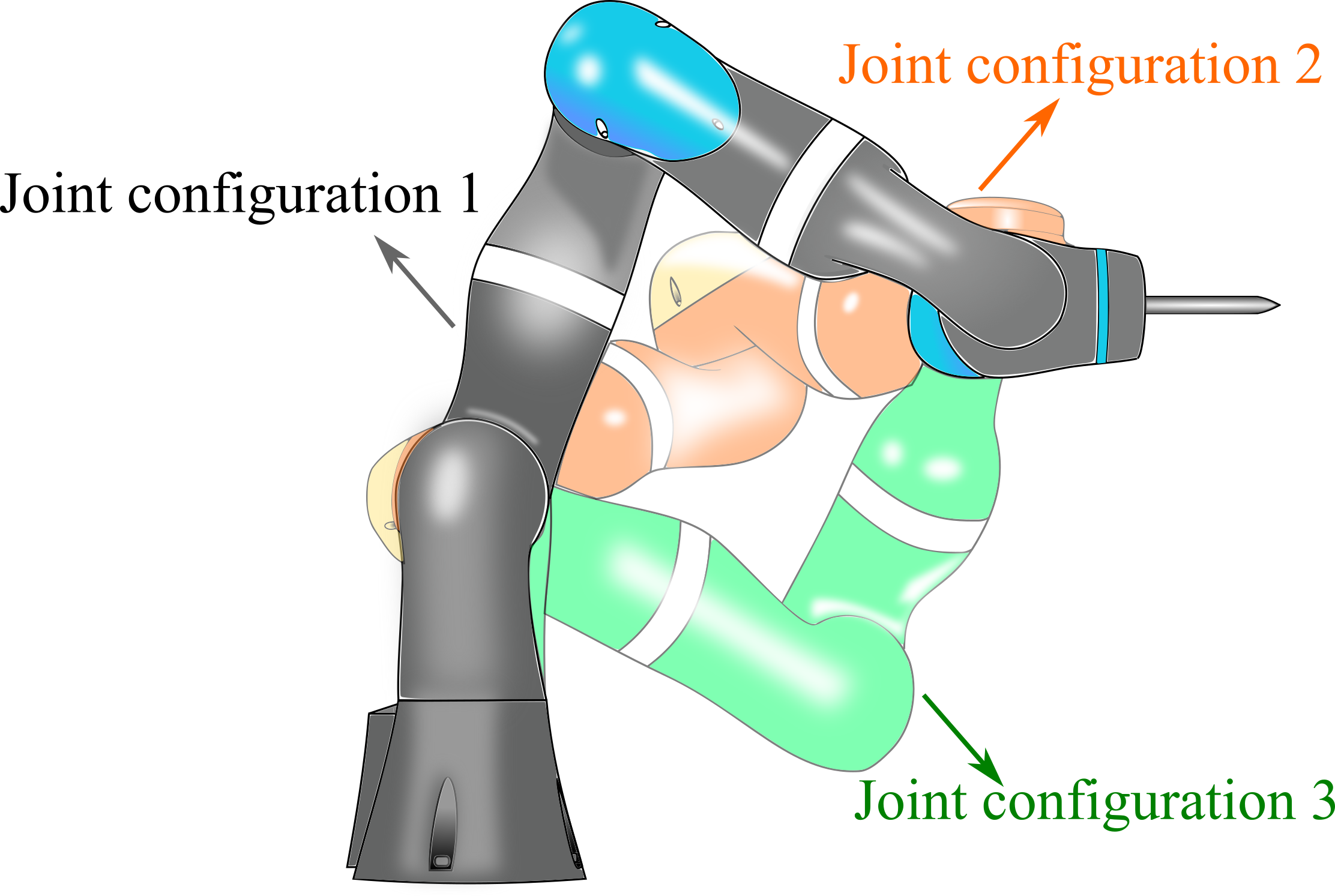}
\caption{KUKA manipulator with a redundant DoF is able to reach each pose via different joint configurations.}
\label{redundant-pose-reach-fig}			% label for the whole figure
\end{figure}

The taxonomy of manipulators types is shown in Fig. \ref{hard2soft-robot-range-fig}\cite{george2018control}. As can be seen, the boundary between hard and soft robots is a gray zone in which redundant and continuum robots exist. Robotic arms made of rigid links chained together are called discrete manipulators. Each discrete robot needs a minimum DoF to reach every point with arbitrary orientation within its task space. Robotic arms that have DoFs more than what is strictly needed to cover their tasks space are called redundant manipulators \cite{patel20052}. Examples of redundant robots are KUKA iiwa \cite{kuka-iiwa} and Kinova Gen2 \cite{kinova-gen2}. The motivation for using extra DoFs is to gain the ability to reach every point of the task space with multiple joint configurations. This is particularly useful to operate in limited spaces. The idea is depicted in Fig. \ref{redundant-pose-reach-fig}. As can be seen, the robot is able to reach the same pose via three different joint configurations, in view of its extra DoF. As a result, if one of the configurations is not feasible due to existing obstacles in the robot task space, the desired pose (i.e. position and orientation) is reached via one of the two remaining joint configurations.

Increasing the DoF to connect a large number of short links leads to continuum manipulators which are a subset of a broader class of hyper-redundant manipulators\cite{inoue2001distributed}. The structure of continuum robots bears resemblance to snake's skeleton (Fig. \ref{hard2soft-robot-range-fig}). The high number of short links makes continuum robots highly flexible, but their motion is still somewhat limited. To make manipulators even more flexible, soft robotic arms have been developed in which links and joints are avoided altogether. Instead of metals and hard plastics, soft robots are commonly made of flexible material \cite{rus2015design} that are safer for applications involving close contact with humans. Soft robots have the potential to insert human's body in a noninvasive manner for performing operations in hard-to-reach areas. Examples of soft robots for medical applications are rehabilitation gloves made of water reservoirs and hydraulic pumps for actuation \cite{polygerinos2015soft} or thumb rehabilitation gloves to aid patients in restoring the correct function of their thumbs \cite{maeder2014biologically}.

However, the use cases of soft robots are not limited to medical applications \cite{lee2017soft}. Multiple soft robots have been developed such as an underactuated robotic hand \cite{tan2017simultaneous}, swimming robots \cite{marchese2014autonomous,sfakiotakis2015octopus}, universal robotic gripper based on jamming granular materials \cite{brown2010universal}, etc.
\section{Manipulator control}\label{manip_control_sec}
The primary goal of a manipulator control module is to have the ability to transfer the manipulator EE to a certain pose in a 3D Cartesian space. While the structural properties of manipulators play a major role in their abilities to complete certain tasks, developing solid controllers for them is just as important. Manipulators and in general any robotic system made of multiple joints (e.g. humanoid robots) are usually controlled using inverse kinematics (IK) \cite{kofinas2015complete,5775598, 6309496,villalobos2022alternative}. The IK module computes appropriate values for manipulator joints such that its EE is transferred to a desired target pose. In addition to computing joints values using IK, it is necessary to compute net reaction forces and net moments for each joint of the robot which is accomplished using inverse dynamics \cite{delp2007opensim}.

After solving the IK problem of a given robot, it is possible to move the robot EE to any desired pose within its workspace by setting joint angles to the values dictated by the IK solution. However, solving the IK problem can be tedious and challenging for complex robots which has encouraged researchers to utilize ML for robot control \cite{wei2003new,csiszar2017solving}. The rest of this paper is devoted to reviewing ML application to manipulator control. Before that, fundamental concepts of RL is reviewed.
\section{Reinforcement learning primer}\label{rl-primer-sec}
RL is a subfield of ML in which an intelligent agent learns to take appropriate actions based on observed states such that a certain utility function is maximized. An RL problem is a sequential decision making task represented as a Markov decision process (MDP) \cite{sutton2018reinforcement}. Each MDP is a 4-tuple element, i.e., $\{S,A,P,r\}$ where $S$ is the state space, $A$ is the action space, $r$ is the reward function, and $P$ is the environment dynamics model. At each time step $t$, the probability of observing state $s_{t+1}$ given that action $a_t$ is taken at state $s_t$ is determined by the dynamics model $P(s_{t+1}|s_t, a_t)$.

An RL agent learns from direct interaction with the environment. In each episode, starting from an initial state $s_0$, the agent takes multiple actions until the problem goal is achieved or certain number of steps are passed upon which episode is terminated. At each time step $t$, action $a_t \in A$ is selected based on observed state $s_t \in S$ and the agent action selection policy $\pi_{\theta}(a_t|s_t)$ with parameter vector $\theta$. The agent learns $\theta$ to maximize the expected sum of discounted rewards:
\begin{equation}\label{rl_objective_eq}
J(\theta)  =\mathbb{E}_{\tau \sim \pi} \left[ \sum_{t=0}^{\infty}\gamma^t r(s_t, a_t)   \right],
\end{equation}
where $\gamma$ is the discount factor and expectation is computed with respect to a set of all possible trajectories $\tau$ when policy $\pi$ is followed. Each trajectory is a sequence of transitions $\{(s_t, a_t, r_t, s_{t+1}), t=0, 1, ...\}$ where $r_t$ equals to the reward function value $r(s_t, a_t)$. The discount factor enforces the fact that as we get further away from the current time step $t$, we care less about the obtained immediate rewards.
\subsection{Policy gradient}
One of the widely used families of RL methods is policy gradient. The first ever policy gradient method called REINFORCE was proposed by Williams \cite{williams1992simple}. The main idea is to optimize parameters vector $\theta$ of a given stochastic policy $\pi_\theta(a_t|s_t)$ such that the objective function in Equation (\ref{rl_objective_eq}) is maximized. To this end, the current version of the policy is used to interact with the environment in order to gather $N$ episodes each of which is at most $T$ steps long. Using the $N$ collected episodes, the gradient in equation (\ref{rl_objective_eq}) with respect to $\theta$ is approximated as \cite{duan2016benchmarking}:
\begin{equation}\label{pg-obj-eq}
\nabla_{\theta}  J(\theta) \approx \frac{1}{NT} \sum_{i=1}^{N} \sum_{t=0}^{T-1} \nabla_\theta \log \left( \pi_\theta(a_t^i | s_t^i) \right) \left( G_t^i - b_t^i \right),
\end{equation}
where $G_t^i$ is the empirical discounted return at time step $t$ of the ith episode computed as:
\begin{equation*}
G_t^i = \begin{cases}
r(s^i_t, a^i_t), \quad t = T\\
r(s^i_t, a^i_t) + \gamma G^i_{t+1} , \quad t=0, ..., T-1
\end{cases},
\end{equation*}
and $b_t^i$ is a baseline term which only depends on $s_t^i$ and it is used to reduce the policy gradient variance.

Despite being simple, REINFORCE has achieved good results in several problems. However, this method conducts its search in the space of policy parameters. Making small modifications in the parameter space can cause significant changes in the policy behavior, leading to its destruction \cite{achiam2017advanced}. Lowering the learning rate can prevent policy destruction but it also slows down the learning process. Alternatively, it is possible to conduct search in the policy space, rather than the parameter space. Natural policy gradient (NPG) \cite{kakade2001natural} is capable of controlling the amount of change between old policy $\pi_{old}$ and new policy $\pi_{new}$ in successive updates \cite{deisenroth2013survey}. This is achieved by limiting the Kullback–Leibler divergence between $\pi_{old}$ and $\pi_{new}$. NPG has been used in the natural actor-critic (NAC) method \cite{peters2008reinforcement, peters2008natural}. Inspired by NPG, trust region policy optimization (TRPO) \cite{schulman2015trust} has been proposed in which the amount of change between old and new policies is forced to be inside a safe region (the trust region). To propose TRPO, Schulman et al. expressed the expected cost of following policy $\pi(a_t | s_t)$ as:
\begin{eqnarray}
\eta(\pi) = \mathbb{E}_{s_0, a_0, ...} \left[ \sum_{t=0}^{\infty} \gamma^t c(s_t) \right], \label{trpo_eta_pi_q}\\ 
s_0 \sim \rho_0(s_0), \: a_t \sim \pi(a_t | s_t), \: s_{t+1} \sim P(s_{t+1} | s_t, a_t),\nonumber
\end{eqnarray}
where $s_0$ is the initial state sampled from probability distribution $\rho_0(s_0)$ and $c(s_t)$ is the cost function. Kakade and Langford \cite{kakade2002approximately} showed that the expected cost of policy $\tilde{\pi}$ can be expressed as the expected cost of another policy $\pi$:
\begin{eqnarray}
\eta_{\tilde{\pi}} = \eta_{\pi} + \mathbb{E}_{s_0, a_0, s_1, a_1,...} \left[ \sum_{t=0}^{\infty} \gamma^t A_{\pi}(s_t, a_t) \right], \label{trpo_eta_tilde_pi_in_terms_of_A_pi_eq}\\
s_0 \sim \rho_0(s_0), \: a_t \sim \tilde{\pi}(a_t | s_t), \: s_{t+1} \sim P(s_{t+1} | s_t, a_t), \nonumber
\end{eqnarray}
where $\eta_\pi$ and $\eta_{\tilde{\pi}}$ are the expected cost of policies $\pi$ and $\tilde{\pi}$, respectively. Moreover, $A_\pi(s_t, a_t) = Q_\pi(s_t, a_t) - V_\pi(s_t)$ is called the advantage function and it is the difference between action-value function $Q_\pi(s_t, a_t)$ and state-value function $V_\pi(s_t)$ \cite{schulman2015trust}:
\begin{align*}
V(s_t) &= \mathbb{E}_{a_t, s_{t+1}, ...} \left[ \sum_{l=0}^{\infty} \gamma^l r_{t+l} \right],\\
Q(s_t,a_t) &= \mathbb{E}_{s_{t+1}, a_{t+1}, ...} \left[ \sum_{l=0}^{\infty} \gamma^l r_{t+l} \right]
\end{align*}
where $r_{t+l}$ is the immediate reward at time step $t+l$ and $\gamma$ is the discount factor. The expectation is taken over stochastic policy $\pi_\theta$ and dynamics model $P(s_{t+1}|s_t, a_t)$. Using equation (\ref{trpo_eta_tilde_pi_in_terms_of_A_pi_eq}), Schulman et al. formulated TRPO algorithm as a constrained optimization problem to update the policy parameters. Both natural policy gradient and TRPO rely on second order derivatives for policy updates. This leads to heavy computation complexity in complex problems. To address this issue, proximal policy optimization (PPO) algorithm \cite{schulman2017proximal} considers the constraint on old and new policy as part of the objective function. This way the computation complexity of PPO becomes much less than TRPO but the aforementioned constraint is sometimes not satisfied which is negligible. PPO has also been extended to Trust Region-based PPO with Rollback (TR-PPO-RB) \cite{wang2020truly} which improves PPO sample efficiency and performance by proposing (1) a new clipping function to limit the difference between new and old policies and (2) revising the condition for triggering the clipping function.
\subsection{Actor-critic architecture}\label{act-critic-sec}
The actor-critic architecture is popular in the RL community. Several well-known RL methods such as deep deterministic policy gradient (DDPG) \cite{lillicrap2015continuous}, asynchronous advantage actor-critic (A3C) \cite{a3cMnih}, proximal policy optimization (PPO), phasic policy gradient (PPG) \cite{ppgCobbe}, soft actor-critic (SAC) \cite{sacHaarnoja} etc. have adopted this architecture. As the name implies, this architecture consists of actors and critics. The role of the critic is to judge the actions taken by the actor ($\pi_\theta$) via estimating the long term utility of those actions. The actor's duty is updating the policy parameters according to the received feedback from the critic. Training of the critic is accomplished using the transitions $\{s_t, a_t, r_t, s_{t+1}\}$ collected during the interaction of the actor with the environment. Considering the RL objective function in Equation (\ref{rl_objective_eq}), critic can be modeled as either $V(s_t)$ or $Q(s_t, a_t)$ \cite{schulman2015trust}.
\section{Control using reinforcement learning}\label{control-using-rl}
In robotic applications, gathering sufficient amount of training data for running supervised learning is rarely practical in terms of cost and time. To gather training data, a human operator must move the robotic arm manually such that the problem goal is fulfilled. In complex problems, this manual data collection process takes a considerable amount of time which is not desirable. Reinforcement learning methods, on the other hand, do not need supervision. They can learn on their own provided that the learning problem is presented as an MDP with an appropriate reward function. RL methods can be divided into two broad classes: model-free RL (MFRL) and model-based RL (MBRL). In RL setup, the environment dynamics model ($P(s_{t+1}|s_t, a_t)$) is usually not known. This is why MFRL methods perform learning based on collected transitions during interaction with the environment.
\subsection{Model-free methods}
Dexterous manipulation is one of the most challenging RL tasks involving multiple robotic manipulators or fingers that cooperate in object manipulation (e.g. grasping and sliding) \cite{okamura2000overview}. Given the high dimensionality of dexterous manipulation tasks, a large number of transitions is needed for training RL agents. Therefore, collecting sufficient number of training samples on real robots is not feasible. Chen et al. \cite{chen2021randomized} proposed the REDQ method with sample efficiency in mind while achieving a performance close to the model-based methods (section \ref{mbrl-sec}). In that study, the ratio of policy updates to collected sample size was set higher than 1 for sample efficiency. To keep the update steps beneficial, an ensemble of networks expressing Q function was used where the networks were initialized randomly but updated with the same target value (Equation (\ref{redq-q-eq})). The computation of target values was performed using a randomly drawn subset of Q networks to reduce the variance of action-value estimation \cite{chen2021randomized}:
\begin{equation}\label{redq-q-eq}
y = r_t + \gamma \left[ \underset{i \in M}{\operatorname{min}} \: Q'_i(s_{t+1}, \tilde{a}_{t+1}) - \alpha \log \pi(\tilde{a}_{t+1}|s_{t+1}) \right], \quad \tilde{a}_{t+1} \sim \pi(s_{t+1}),
\end{equation}
where $M$ is the number of randomly chosen Q networks and $\alpha$ determines the importance of the logarithm term against the action-value function. The minimum is computed over the soft state-value function of the SAC method. REDQ can be used alongside any off-policy MFRL methods like TD3 \cite{dankwa2019twin}, SAC, and SOP\cite{wang2020striving}.

As an alternative method for sample efficiency, researchers conduct training in simulation \cite{ZhuRlVisumotor,peng2018sim,rusu2017sim}. No matter how complex and accurate, existing simulators still cannot capture the full characteristics of real robots leading to the gap between simulation and reality. Obviously, an agent trained on simulated data fails to act appropriately on a real robot. As a workaround, dynamic randomization\cite{SadeghiL16,tobin2017domain, peng2018sim} can be used to run multiple instances of simulations each of which has different parameter values. As an example, to train the 24-DoF Shadow hand \cite{shadowhand} for object manipulation, Andrychowicz et al. \cite{openai2018learning} utilized dynamic randomization. Multiple instances of Shadow hand simulation with different sets of parameters (e.g. friction coefficients and object's appearance) were run. The simulation was done using the OpenAI Gym benchmark \cite{plappert2018multi, 1606.01540} implemented using MuJoCo  \cite{todorov2012mujoco}. The collected data were used to train OpenAI Five implementation \cite{OpenAI_dota} of PPO agent, which was later tested on a real robot setup. Apart from filling the gap between simulation and reality, running multiple simulation instances in parallel accelerates the data collection process \cite{nguyen2019review}.

Given the success of DDPG in various RL problems with continuous state and action spaces, it has been utilized to learn the control policy for an UR5 manipulator \cite{9164440}. A reward function based on a  manipulability index was proposed. The manipulability index reflects the robot ability to move in a specific direction and it is computed based on mapping robot's operational space to joint space. The manipulability index has been reported to be a better measure of dexterity, as compared with the condition number or minimum singular value \cite{patel2015manipulator}. However, reward engineering is not always easy. As such, in addition to actual interaction with the environment, Vecerik et al. \cite{vecerik2017leveraging} utilized human demonstrations to aid DDPG training in learning complex manipulation tasks. The demonstrations were conducted using a kinesthetically controlled robot. Similarly, inspired by \cite{peters2008reinforcement}, Rajeswaran et al. \cite{rajeswaran2017learning} utilized demonstration to improve sample efficiency of natural policy gradient \cite{kakade2001natural, peters2007machine, rajeswaran2017towards}. The resulting approach is called data-augmented policy gradient (DAPG) in which the policy parameters are initialized by behavior cloning (BC) \cite{bojarski2016end,pomerleau1988alvinn} via solving the following maximum-likelihood problem \cite{rajeswaran2017learning}:
\begin{equation}
\underset{\theta}{\operatorname{max}} \sum_{(s,a) \in \rho_D} \ln \pi_\theta(a|s),
\end{equation}
where $\rho_D$ is the set of demonstration samples. In addition to using BC, DAPG modifies the standard policy gradient objective function (Equation (\ref{pg-obj-eq})) by adding an extra term for combining RL with imitation learning:
\begin{equation}\label{dapg-obj-eq}
\nabla_{\theta}  J(\theta) \approx \frac{1}{NT} \sum_{i=1}^{N} \sum_{t=0}^{T-1} \nabla_\theta \log \left( \pi_\theta(a_t^i | s_t^i) \right) \left( G_t^i - b_t^i \right)+ \sum_{(s,a) \in \rho_D} \nabla_\theta \ln \pi_\theta(a|s)w(s,a),
\end{equation}
where $w(s,a)$ is a weighting function that decays as the training unfolds. The reason for the additional term in Equation (\ref{dapg-obj-eq}) is that while BC is useful, policies initialized by this method are not necessarily successful due to distribution shifts between policy's states and demonstrated ones \cite{ross2011reduction}. While BC helps with policy initialization, it fails to capture correct sequence of behaviors such as reaching, grasping, and moving an object. This requirement is enforced by the additional term in Equation (\ref{dapg-obj-eq}).

The power of supervised learning has also been exploited to tackle high dimensional manipulation tasks. Levine et al. \cite{levine2016end} proposed a guided policy search method to train the vision and control modules of an PR2 robot to carry out complex tasks of cap screwing on a bottle. Their method consists of outer and inner loops. In the outer loop, multiple linear Gaussian controllers $p_i(u_t|x_t)$ are run on the actual robot to collect trajectory samples $\{\tau_i^j\}$ corresponding to different initial states. Dynamics models $p_i(x_{t+1}|x_t, u_t)$ are fit to the collected trajectories and are used to improve $p_i(u_t|x_t)$. The inner loop alternates between optimizing $p_i(\tau)$ and optimizing the policy so that it is matched against the trajectory distributions. The alternating optimization is implemented using the Bregman alternating direction method of multiplier (BADMM) algorithm \cite{wang2014bregman}. The policy is trained to predict action sequence for each trajectory using only observation $o_t$ instead of full state $x_t$. The motivation for using $o_t$ is that the policy can operate directly on raw  observations (i.e. images) at inference time.
\subsection{Model-based methods}\label{mbrl-sec}
MBRL methods consist of two phases: interaction with the environment and simulation. During the interaction phase, transitions are collected just like the MFRL method. The gathered data are used to learn the dynamics model. During the simulation phase, the learned dynamics are used for the next state prediction in successive steps without actual interaction with the environment. The simulated samples are then used for updating the action selection policy.

The dynamics model can be used for simulated data generation, planning, or value function expansion. State-only imitation learning (SOIL) \cite{radosavovic2021state} is one of the MBRL methods that uses dynamics models for simulated data generation. SOIL collects the state sequences observed in the environment. It relies on an inverse dynamics model to predict the most likely action sequence that may have led to those state sequences. The evaluation of SOIL has been completed in simulation. Valencia et al. \cite{valencia2023} took a step further and investigated the possibility of using MBRL on real robots. They utilized an ensemble of probabilistic networks and mixture of Gaussian distributions to generate simulated samples with better quality and capture dynamics model uncertainty. Twin-delayed DDPG (TD3) was used as the RL agent. An ensemble of probabilistic networks \cite{chuaPets} was utilized by the MBPO method \cite{janner2019trust} to deal with the two sources of the dynamics model error. The first source is lack of knowledge about unseen regions of the state-action space leading to generalization errors. The second source is the changing behavior of the policy, owing to its updates which causes distribution shift. MBPO aims to manage these error sources by keeping the simulation horizon short and handling uncertainty using probabilistic networks. To this end, multiple transitions are randomly drawn from the pool of collected samples and expanded for $K$ simulated time steps where $K$ is the simulation horizon. The simulated and real training samples are used to update the policy using the SAC method.

Different approaches have been taken to implement planning-based MBRL methods. Omer et al. \cite{9429677} fused SAC and model predictive control (MPC) to improve data efficiency of the MPC-MFRL approach \cite{hong2019model}. Generative adversarial tree search (GATS) \cite{azizzadenesheli2018sample} used generative adversarial networks (GAN) \cite{goodfellow2014generative} for dynamics model representation and Monte Carlo tree search \cite{coulom2006efficient} for planning. PILCO \cite{deisenroth2011pilco} was used to learn legged locomotion \cite{deisenroth2012toward} and control of a low-cost manipulator \cite{deisenroth2011learning}. PILCO is extremely powerful in capturing the uncertainty of the environment dynamics due to using Gaussian process (GP) \cite{williams2006gaussian} as its dynamics model. However, GP is a non-parametric method with $O(n^3)$ complexity which scales with number of samples $n$. To keep the computation complexity bounded, probabilistic ensembles with trajectory sampling (PETS) \cite{chuaPets} utilized an ensemble of probabilistic neural networks to capture epistemic and aleatoric uncertainties simultaneously. During the agent interaction with the environment, the state-action space is only partially observed, which is the cause of epistemic uncertainty. Aleatoric uncertainty is due to the inherent noise in the environment dynamics. PETS and some other MBRL methods \cite{hafner2019learning,wang2019exploring} used the cross entropy method \cite{rubinstein1999cross} for online planning. However, CEM is not scalable to high dimensional tasks of vision-based RL \cite{kotb2023sample}. In these problems, the visual states are mapped onto a latent space using DNNs. The reason that CEM is not scalable to high dimensional tasks is that it randomly generates actions sequences and executes the first action of the sequence with the highest expected reward. This is not an efficient strategy \cite{wang2019exploring, nagabandi2018neural}. Moreover, CEM solely attempts to maximize the extrinsic reward (obtained from the environment) of the sampled action sequence. As such, CEM lacks exploration which is crucial to successful learning in problems with sparse rewards and hard-to-explore state space. These drawbacks of CEM have prevented MBRL methods from reaching asymptotic performance of MFRL methods \cite{chuaPets}. The exploration issue can be addressed using intrinsic rewards based on the prediction error of the next latent space \cite{pathak2017curiosity,oudeyer2007intrinsic,burda2018large}. While intrinsic motivation based on curiosity has been used in multiple MFRL methods \cite{pathak2017curiosity, burda2018large, houthooft2016vime,mohamed2015variational}, it has seen limited application in MBRL \cite{sekar2020planning}. This stems from the fact that ground truth for the next observation in the latent space is not available during online planning. To address this issue, Plan2Explore \cite{sekar2020planning} leveraged the disagreement between an ensemble of dynamics models as an estimate for the intrinsic reward. However, training multiple dynamics models is computationally intensive. This leads to the emergence of the curiosity CEM \cite{kotb2023sample} approach which achieves exploration via curiosity by estimating intrinsic reward offline during training.

The value function expansion approach of MBRL is usually paired with actor-critic methods. The actor parameters are updated such that the utility it receives from the critic element is maximized. As an example, the DDPG actor is trained by minimizing the following loss function over a batch of $N$ transitions $\{(s_t, a_t, r_t, s_{t+1})^{(i)}, i=1, ..., N\}$: 
\begin{align*}
\theta_{new} \leftarrow \underset{\theta}{\operatorname{argmin}} -\frac{1}{N}\sum_{i=1}^{N} Q(s_t^{(i)}, \pi_\theta(s_t^{(i)})).
\end{align*}
where $\theta_{new}$ is the vector of policy (actor) updated parameters. The critic parameters are updated such that the following mean squared error (MSE) is minimized:
\begin{equation*}
\phi_{new} \leftarrow \underset{\phi}{\operatorname{argmin}} \frac{1}{N}\sum_{i=1}^{N} \left[ Q_\phi(s_t^{(i)}, a_t^{(i)}) - y^{(i)} \right]^2,
\end{equation*}
where $\phi$ is the critic parameters vector. Considering that the true $Q(s_t, a_t)$ values are unknown, the target values for training DDPG critic are estimated as:
\begin{equation}\label{ddpg-critic-target-eq}
y^{(i)} = r_t^{(i)} + \gamma Q'(s_{t+1}^{(i)}, \pi'_\theta(s_{t+1}^{(i)})),
\end{equation}
where $Q'(.,.)$ and $\pi'(.)$ are called the target networks that slowly track the changes of $Q(.,.)$ and $\pi(.)$ parameters via Polyak averaging \cite{lillicrap2015continuous}.  Value expansion methods such as Model-based value expansion (MVE) \cite{mveFeinberg} and Stochastic ensemble value expansion (STEVE) \cite{buckman2018sample} expand the one-step prediction $y^{(i)}$ of Equation (\ref{ddpg-critic-target-eq}) to multiple steps in an attempt to make it more accurate accelerating the learning process. The expansion is achieved via simulation based on the actor and the learned dynamics model. Contrary to MVE that expresses dynamics model with a single neural network, STEVE utilizes an ensemble of neural networks to capture the uncertainty of the learned dynamics model.
% loca regret 54
% mve, steve
% \cite{9195341}: ensemble of dyn models, tested on simulated Baxter robot in ROS
\subsection{Hierarchical RL}\label{hrl-sec}
The search space of complex robotic tasks is continuous and high dimensional which makes achieving effective exploration challenging. Such tasks require long sequences of successive transitions $\{(s_t, a_t, r_t, s_{t+1}), t=1, ..., T\}$ (known as rollouts) before the RL agent can reach its goal \cite{8629360}. At the start of learning, the agent policy is not good enough to take appropriate actions. As a result, reaching key points in the task state space is unlikely, which slows down the learning process. However, Hierarchical RL (HRL) methods can exploit highly structured nature of robotics problems \cite{daniel2016hierarchical} to accelerate learning. Some HRL methods utilize multiple skills \cite{agarwal2018deep,agarwal2019model} as sub-policies and a gating policy to execute appropriate skills based on observed states. Hierarchical relative entropy policy search (HiREPS) \cite{daniel2016hierarchical} adopts such an approach and uses softmax function as gating policy and dynamic movement primitives (DMP) \cite{ijspeert2002movement, schaal2006dynamic,ijspeert2013dynamical} as sub-policies. Owing to the smooth and flexible trajectory generation of DMPs, HiREPS can learn real robot manipulation tasks. HiREPS casts the HRL problem into a convex optimization problem and limits the difference between policies from successive training iterations to avoid premature convergence. While HiREPS attempts to keep the sub-policies versatile, it has no mechanism to enforce nonzero selection probabilities for all of sub-policies. Therefore, sometimes HiREPS sticks to updating just one sub-policy. Moreover, HiREPS lacks the feature to set hyperparameters for gating and sub-policies independently which hinders learning. End et al. tried to address these issues \cite{end2017layered}

\begin{figure}[] 
\centering
\includegraphics[width=0.7\textwidth]{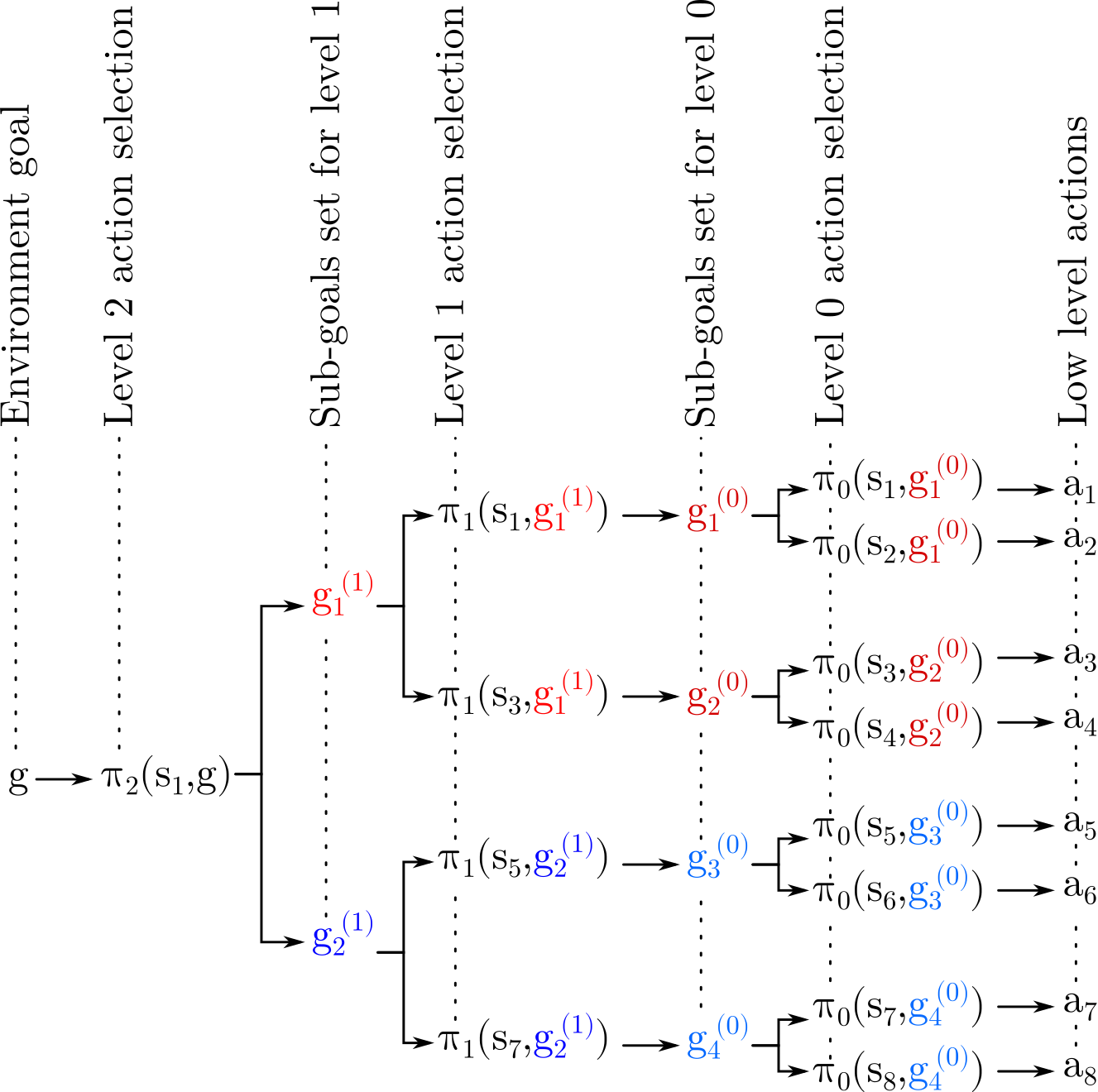}
\caption{Illustration of how HAC hierarchical policy with three layers and time scale two interacts with the environment: time scale is number of time steps ith layer can perform action selection before returning control to its upper layer for action selection.}
\label{temporal-abstraction-fig}			% label for the whole figure
\end{figure}

Another category of HRL methods is based on temporal abstraction which breaks long rollouts into multiple shorter ones by choosing appropriate intermediate goals. Reaching these goals leads to the realization of the original goal. Multi-goal RL methods \cite{plappert2018multigoal} are capable of achieving multiple goals offering better generalization and possibility for curriculum learning \cite{beyret2019dot}. Hierarchical actor-critic (HAC) \cite{levy2017hierarchical,levy2017learning} is one of the first temporal abstraction methods capable of tackling multi-goal manipulation problems. HAC utilizes DDPG for policy learning, universal values function approximators (UVFAs) \cite{schaul2015universal} for multi-goal support and hindsight experience replay (HER) \cite{andrychowicz2017hindsight} to handle sparse rewards (Section \ref{reward-fcn-design-sec}). UVFA is a generalized value function $Q(s,a,g)$ that performs value estimation based on state $s$, action $a$, and goal $g$. As an example, in a robotic manipulation task, goal can be the desired pose for the robot EE. The main idea of HAC is to use a multi-layer hierarchical policy. A sample hierarchical policy with three layers is depicted in Fig. \ref{temporal-abstraction-fig}. At the start of the rollout, based on state $s_1$ and the environment goal $g$, the top layer policy $\pi_2$ chooses action $g_1^{(1)}$ as the goal of the middle layer policy $\pi_1$ to set goal $g_1^{(0)}$ for the bottom layer policy $\pi_0$. After action selection for two time steps, $\pi_0$ returns the control to $\pi_1$ which selects action $g_2^{(0)}$ based on state $s_3$ and goal $g_1^{(1)}$. Once again $\pi_0$ interacts with the environment by selecting actions $a_3$ and $a_4$ after which the control is returned to $\pi_2$ to select another goal $g_2^{(1)}$. This trend continues until one rollout is complete. The advantage of temporal abstraction methods like HAC is that exploration is performed at all layers of the policy simultaneously which accelerates learning and increases the chance of visiting key states in complex environments.

The drawback of HAC is that the time steps assigned to each of the policies at different layers is a fixed value. Referring to Fig. \ref{temporal-abstraction-fig}, the time scale is two, which means at each layer $i$, policy $\pi_i$ is allowed to perform action selection twice in order to reach the goal $g_*^{(i)}$ assigned to it. Setting the right time scale has a direct impact on the learning progress of the HAC method. This issue has been addressed in HiPPO \cite{li2019sub} which treats the time scale as a random variable sampled from a specific distribution. The randomization technique used by HiPPO leads to more flexible and faster learning. HiPPO leverages a two-layer hierarchical extension of PPO to control the behavior of the bottom layer according to a latent variable set by the top layer. The value of the latent variable is similar to the number of sub-policies in methods like HiREPS. Using the latent variable, HiPPO can achieve easy to hard learning. Similar curriculum learning has been achieved by options of interest (IOC) method \cite{khetarpal2020options}, which controls its bottom layer according to the option index set by the top policy. Another drawback of HAC is the mismatch between transitions %$\{(s_t||g^{(i+1)}_{t'}, g_t^{i}, r_t, s_{t+\zeta}||g_{t'}^{(i+1)}), t=1, ..., T\}$
stored in the replay buffer of layer $i+1$ and the behavior of layer $i$. Considering that the policy layers are updated in parallel, the old actions that have been chosen by layer $i+1$ do not necessarily reflect the behavior of the updated policy at layer $i$. This mismatch can slow down learning. To address this issue, another HRL method called HIRO \cite{nachum2018data} proposed off-policy correction. HIRO considers a two-layer policy and revises the actions chosen by the top policy such that the likelihood of the transitions generated by the bottom policy is maximized.

Considering that curiosity has a hierarchical nature \cite{pezzulo2018hierarchical}, HAC has been extended to curious hierarchical actor critic (CHAC) \cite{roder2020curious} which offers a hierarchical curiosity mechanism to handle sparse rewards. HAC was accompanied by an additional reward signal encouraging curiosity. In addition to HER, curiosity is another method enabling learning from sparse rewards. The main idea is that the RL agent receives additional rewards for getting surprised \cite{pathak2017curiosity, hafez2017curiosity} which acts as an incentive for exploring unseen states. The surprise element can be expressed as the agent's dynamics model prediction error \cite{friston2011action}. Therefore, CHAC considers a dynamics model $f_w: S \times A \rightarrow S$ for each policy layer. Model $f_w$ estimates the next state $s_{t+1}$ given current state $s_t$ and action $a_t$. For each layer $i$, the intrinsic reward based on curiosity at time step $t$ is expressed as:
\begin{equation*}
r_{c,t}^{(i)} = \frac{\left( s^{(i)}_{t+1} - \hat{s}^{(i)}_{t+1} \right)^2}{2},
\end{equation*}
Using $r_{c,t}^{(i)}$, the total reward for layer $i$ at time step $t$ is computed as:
\begin{equation*}
r_t^{(i)} = \alpha r_t^{(i)} + (1-\alpha) r_{c,t}^{(i)},
\end{equation*}
where $r_t^{(i)}$ is the extrinsic reward (obtained from the environment) and $\alpha$ is coefficient in range $(0,1)$.

Yet another extension of HAC is called twin delayed HAC (TDHAC) \cite{anca2021twin}. This method utilizes a two-layer policy updated using the TD3 method. Moreover, compared with HAC, the replay buffer size of the top layer is reduced according to the time scale variable to avoid performance degradation due to an unnecessarily large buffer \cite{zhang2017deeper}. The applied modifications lead to successful learning of a pick-and-place task involving object interactions, which has not been investigated in the original HAC paper. Due to space limit, the rest of the HRL methods are summarized in Table \ref{hrl-summary-table}.

% To reduce left and right margins of table cells, "\hspace{?pt}}" command in 
%https://tex.stackexchange.com/questions/439824/tables-how-to-control-left-and-right-cell-margins
% has been used

\begin{table}[]
\centering
\caption{Summary of HRL methods}
\label{hrl-summary-table}
\begin{adjustbox}{width=1\textwidth}
\small
\begin{tabular}{|c|p{0.48\textwidth}|c|c|c|c|}
\hline
\begin{tabular}{@{}c@{}}Method,\\Year\end{tabular} & Approach & \begin{tabular}{@{}c@{}}Sparse\\reward\end{tabular} & Simulator & \begin{tabular}{@{}c@{}}Real robot\\experiments\end{tabular} & \begin{tabular}{@{}c@{}}Safety\\aware\end{tabular} \\
\hline
\begin{tabular}{@{}c@{}}AHRM\cite{tao2022multi},\\2022\end{tabular}    & Guiding PPO agent by adaptive hierarchical reward mechanism; learning task objectives according to their priority hierarchy         & -                  & \begin{tabular}{@{}c@{}}V-REP\\\cite{rohmer2013v}\end{tabular}   & \begin{tabular}{@{}c@{}}JACO arm\\\cite{campeau2019kinova}\end{tabular}  & -     \\
\hline
\begin{tabular}{@{}c@{}}PG-H-RL\cite{jung2022physics},\\2022\end{tabular}    & Improving TD3 agent efficiency using physics-guided hierarchical reward mechanism & - & V-REP  & \begin{tabular}{@{}c@{}}MICO arm\\\cite{campeau2019kinova}\end{tabular}  & -         \\
\hline
\cite{pinsler2018sample}, 2018 & Hierarchical policy: {upper layer: contextual GP-UCB acquisition function \cite{srinivas2009gaussian, krause2011contextual}, lower layer: parameterized Gaussian}, lower layer learner: REPS \cite{peters2010relative} & - & Unknown & - & -\\
\hline
\begin{tabular}{@{}c@{}}HRL-DCS\cite{ren2022research},\\2022\end{tabular} & Dual arm control, two-layer policy with different time scales using LSTM \cite{hochreiter1997long}, upper layer provides goal and parameterized reward for lower layer, potential field obstacle avoidance  & - & Unknown & \begin{tabular}{@{}c@{}}Dual arms\\with redundant\\DoFs\end{tabular} & -\\
\hline
\cite{de2022learning}, 2022 & Dual arm object manipulation, discretized manipulation spaces, DQN agent \cite{mnih2015human}   & - & Unknown & \begin{tabular}{@{}c@{}}Dual arms\\with 5 DoFs\end{tabular} & -\\
\hline
\begin{tabular}{@{}c@{}}SAC-LSP\cite{haarnoja2018latent},\\2018\end{tabular} & Based on maximum entropy RL \cite{todorov2006linearly, aghasadeghi2011maximum}, optimal control problem transformed to an inference problem using probabilistic graphical model \cite{toussaint2009robot} & \checkmark & OpenAI Gym & - & -\\
\hline
\begin{tabular}{@{}c@{}}HIDIO\cite{zhang2021hierarchical},\\2021\end{tabular} & Two layer policy, top layer outputs vectors $u \in [-1,1]^D$ as task-agnostic options, bottom layer takes primitive actions according to received option. Both layers are trained using SAC & \checkmark & \begin{tabular}{@{}c@{}}MuJoCo,\\SocialRobot\\\cite{socialRobot}\end{tabular} & - & -\\
\hline
\begin{tabular}{@{}c@{}}DHRL\cite{lee2022dhrl},\\2022\end{tabular} & The relation between the top and bottom layer horizons is broken to let the two layers operate at horizons that suit them best. & \checkmark & MuJoCo & - & -\\
\hline
\begin{tabular}{@{}c@{}}PAHRL\cite{gieselmann2021planning},\\2021\end{tabular}  & Hybrid planning/RL method based on DHAC \cite{bellemare2017distributional}, the task MDP decomposed to multiple MDPs with shorter horizons & \checkmark & \begin{tabular}{@{}c@{}}PyBullet\\\cite{coumans2016pybullet}\end{tabular} & -  & - \\
\hline
\begin{tabular}{@{}c@{}}DSG\cite{bagaria2021skill},\\2021\end{tabular} & Uses extension of skill chaining \cite{konidaris2009skill} called deep skill chaining \cite{bagaria2020option, bagaria2021robustly} for building chain of options, the options are used to navigate between regions discovered by the proposed graph expansion method (RRT \cite{lavalle1998rapidly} extended to RL) & - & MuJoCo & -  & - \\
\hline
\end{tabular}
\end{adjustbox}
\end{table}

\subsection{RL challenges in robot control}
Despite the fact that RL does not need labeled training data, applying it to complex robot control problems is not straightforward. Several challenges must be addressed before RL can be used effectively for robot control.
\subsubsection{Enforcing robot and user safety}\label{enf-robot-user-safe-sec}
In the RL literature, it is common practice to make the agent's actions noisy to avoid greedy exploitation of what it has learned, in order to force exploring unseen regions of the search space. Some of the exploration methods are adding Gaussian noise or the Ornstein–Uhlenbeck process \cite{uhlenbeck1930theory} value to the deterministic action or perturbing the parameters of the policy via parameter space noise exploration \cite{plappert2017parameter}. Despite being effective, the aforementioned methods are not safe to execute on real robots since they often result in jerky or sudden movement patterns of the robot joints, resulting in potential harm to the robot and/or user. Therefore, in robot applications, exploration must be effective and safe such as generalized state dependent exploration (gSDE) \cite{raffin2022smooth} which performs exploration based on the current state of the robot to avoid dangerous or jerky motions.

In addition to safe exploration, initializing policy parameters based on appropriate demonstrations can also enforce safety. Grasping is one of the core skills in various robotic tasks. Attention-based deep RL has been used to learn safe grasping policies from demonstration using a Next-Best-Region attention module \cite{10059127}. Alternatively, safety can be enforced by projecting policy output (i.e. selected action) onto a safe action space \cite{emam2021safe} using control barrier functions (CBFs) \cite{7759067}. A CBF $B(x):C \rightarrow \mathbb{R}$ is a non-negative function defined to enforce a specific constraint \cite{rauscher2016constrained}. Safe exploration is not limited to RL \cite{gu2023humancentered}. Any ML method must be safe if it is to be deployed in the real world \cite{turchetta2019safe, baumann2021gosafe, kaushik2022safeapt}. For further details, the interested reader is referred to the recent survey on safe RL \cite{gu2022review}.
\subsubsection{Noisy insufficient training data}
It is always challenging and costly to gather training data using real robots. That is why the issue of data efficiency has been actively researched in the RL community, leading to MBRL approaches (Section \ref{mbrl-sec}) as well as sophisticated robotic simulators. For human-robot collaboration applications, simulations must be accompanied by some sort of human interaction. Virtual/augmented reality is one possible approach to address the human-in-the-loop requirement \cite{lee2022towards}.
Addressing training data scarcity is only part of the solution toward implementing safe RL on real robots. In particular, even collected data can be noisy and misleading due to erroneous intentional/unintentional safety signals provided by human operators. Adversarial training is useful to learn robust policies \cite{meng2023integrating} but it needs further investigation \cite{lechner2021adversarial}.
\subsubsection{Transparent behavior}
To assess safety of RL methods, they must be transparent to the user. The term explainable AI (XAI) is often encountered in applied DL literature \cite{kamath2021explainable, arrieta2020explainable} such as medical diagnosis \cite{moreno2023data, aelgani2023local, mukhtorov2023endoscopic, kolarik2023explainability}, autonomous navigation \cite{madhav2022explainable, onyekpe2022explainable, atakishiyev2021explainable}, and manufacturing \cite{yoo2021explainable, soldatos2021trusted, senoner2022using, ahmed2022artificial}. However, given that modern RL methods heavily rely on complex (possibly deep) neural networks, XAI has become a concern in RL as well \cite{he2021explainable}.
\subsubsection{Reward function design}\label{reward-fcn-design-sec}
\begin{figure}[] % use figure* instead of figure if you want to span subfigures across two columns
\centering
\subfloat[]{\includegraphics[width=0.3\textwidth]{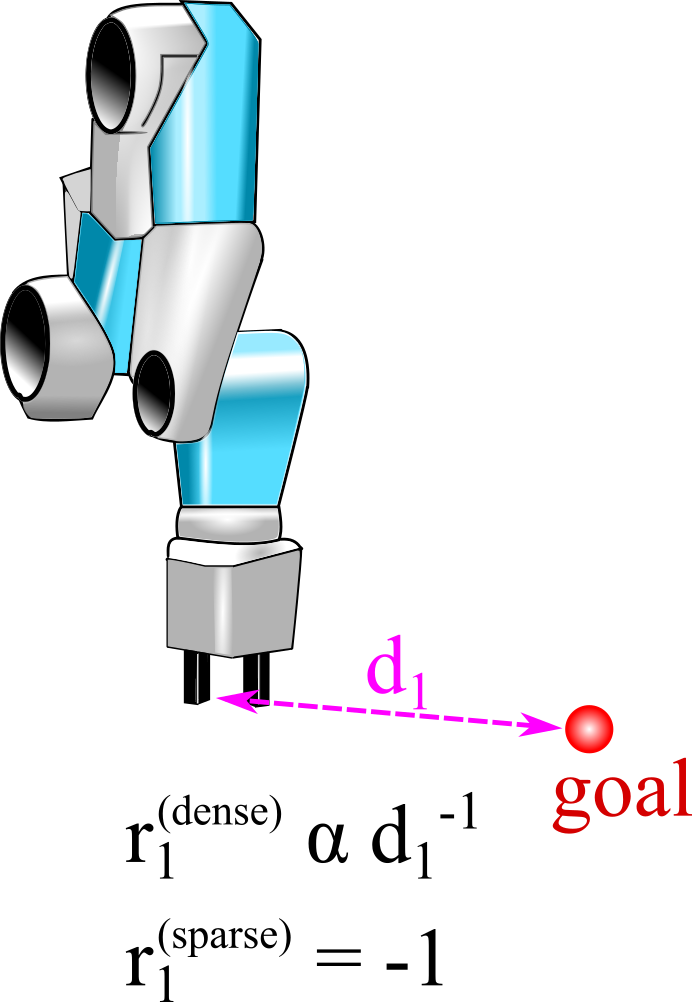}
\label{reward1-fig}}
\subfloat[]{\includegraphics[width=0.3\textwidth]{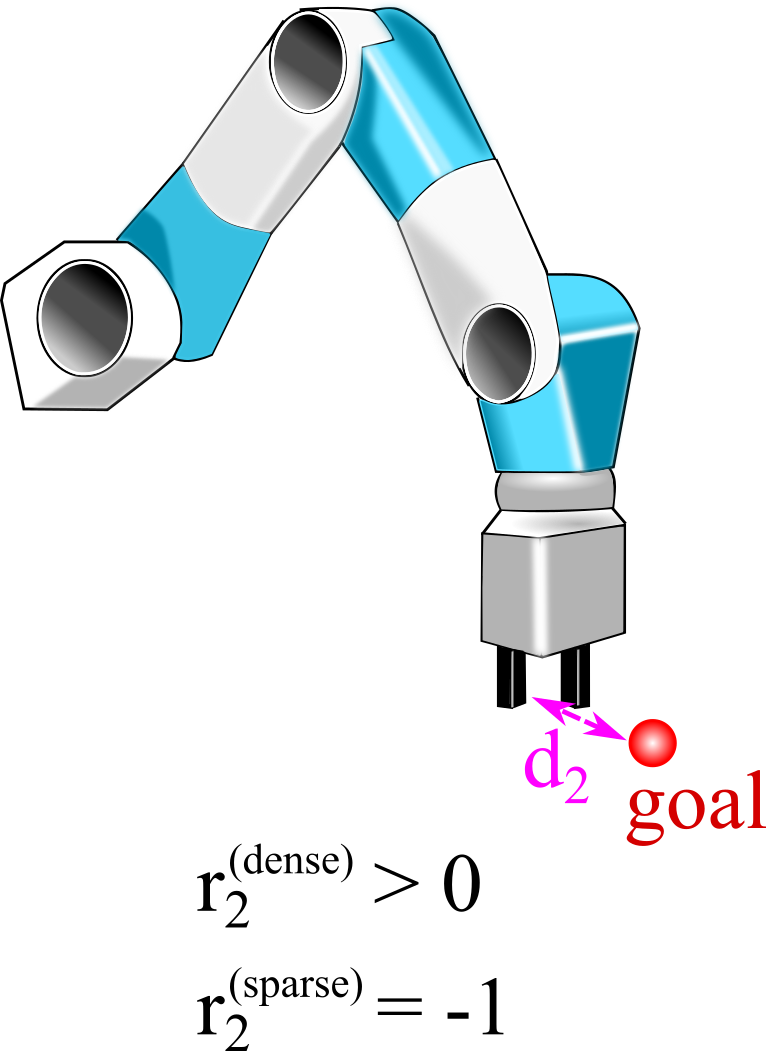}
\label{reward2-fig}}
\subfloat[]{\includegraphics[width=0.3\textwidth]{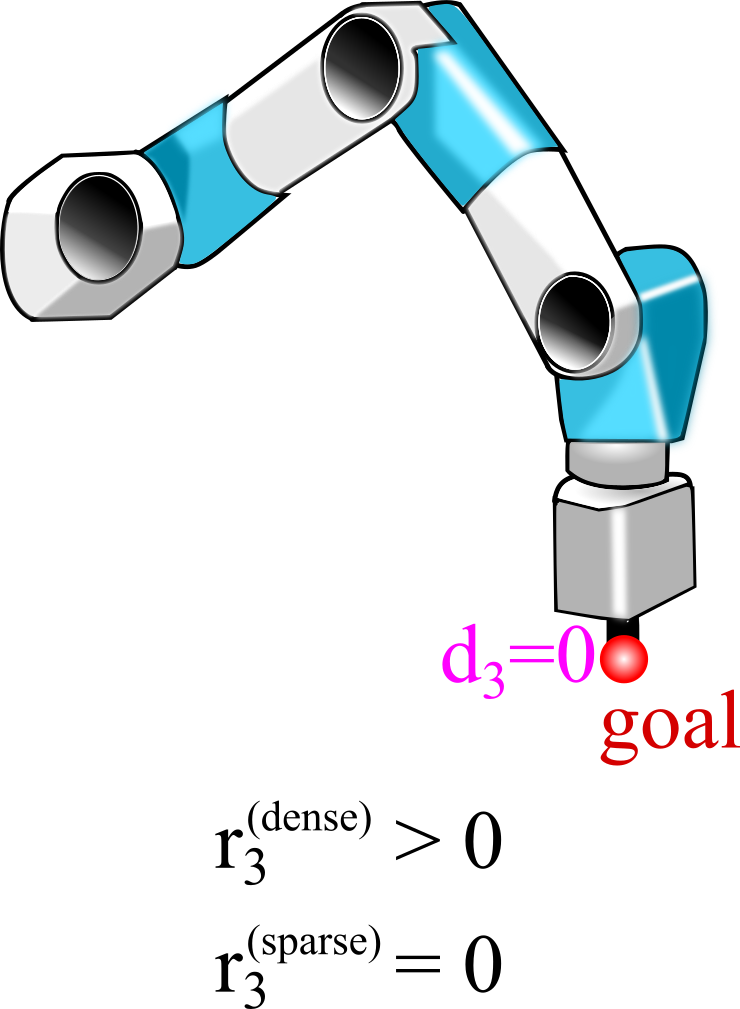}%
\label{reward3-fig}}
\caption{Illustration of dense and sparse reward: (a) distance between goal (red point) and robot EE is $d_1>0$, (b) EE has gotten closer to the goal ($d_2 < d_1$), (c) EE has reached the goal ($d_3 = 0$).}
\label{dense-sparse-reward-fig}			% label for the whole figure
\end{figure}

In the RL literature, the reward values are either dense or sparse. A dense reward increases as the RL agent approaches its goal. On the other hand, a sparse reward is binary; it is always $-1$ unless the agent achieves its goal which yields a reward of zero. The difference between sparse and dense rewards is depicted in Fig. \ref{dense-sparse-reward-fig}. Initially, the distance of the robot EE  to the goal is $d_1$, yielding dense reward $r_1^{(dense)} > 0$ which is inversely proportional to $d_1$. Considering that the goal has not been reached in Fig. \ref{reward1-fig}, sparse reward $r_1^{(sparse)}$ is $-1$. In Fig. \ref{reward2-fig}, similar argument holds yielding sparse reward $r_2^{(sparse)}=-1$ but dense reward $r_2^{(dense)}$ increases owing to reduction of distance between the EE and the goal ($d_2 < d_1$). Finally, in Fig. \ref{reward3-fig}, the robot EE reaches the goal, receiving sparse reward $r_3^{(sparse)}=0$ and higher dense reward ($r_3^{(dense)} > r_2^{(dense)} > r_1^{(dense)}$).

As can be seen, dense rewards guide the RL agent toward achieving their goals but are harder to design for complex problems. Sparse rewards are easily definable for complex problems but RL agents have a tough time learning anything useful from them. This is because the agent does not know how to the goal it is. To remedy this issue, HER has been proposed. Following promising results of HER, multiple challenging robotic tasks were developed in simulation \cite{plappert2018multigoal} motivating further research on HER for tackling sparse reward RL \cite{fang2019curriculum, silver2018residual,li2020generalized,ren2019exploration}.

\subsubsection{Valid action generation}
RL agents may produce actions that are not attainable by the IK module. Assume the action is a 6D pose of the robot EE i.e. $(x,y,z)$ position and (roll, pitch, yaw) orientation. Considering that these six variables are correlated by the robot structural limitations, setting lower and upper bounds for them does not necessarily enforce validity of the agent's actions. Domain knowledge in the form of expert's demonstrations can be used to initialize agent's policy in way that selected actions are valid. As an example, DMP parameters can be adjusted to reflect the expert's demonstration leading to smooth and safe robot trajectories.

\section{Deep learning integration with manipulators}\label{dl-manip-sec}
DL can handle high dimensional data with ease which makes it the method of choice for addressing different challenges of manipulation problems \cite{james2022q,zhu2022viola, julian2020efficient, lee2022spend, jia2020vision, akkaya2019solving, kalashnikov2018scalable}. DL has been used to model dynamics of a 6-DoF manipulator \cite{liang2018robot}. A multilayer perceptron (MLP) with four hidden layers was used. The network inputs were six-dimensional (corresponding to 6-DoF) position vector $q$, velocity vector $\dot{q}$, and acceleration vector $\ddot{q}$. The network output was the corresponding torque vector $\mathcal{T}$ computed as \cite{liang2018robot}:
\begin{equation*}
\mathcal{T} = M(q)\ddot{q}+b(q,\dot{q})+g(q)+f(\dot{q}),
\end{equation*}
where $M(q)$ is the inertial term related to the joint positions, $b(q,\dot{q})$ is the Coriolis \cite{thomasCoriolis} and centrifugal forces, $g(q)$ is gravity, and $f(\dot{q})$ represents friction. In addition to dynamics modeling, DL can also be used to transfer expert knowledge to robotic arms. A notable example is the time-contrastive network (TCN) \cite{sermanet2017time, sermanet2018time} which learns manipulation tasks like pouring liquid into a glass from unlabeled videos containing human demonstrations. TCN is viewpoint-invariant so it is able to observe a specific behavior, e.g. from a third person's view and mimic it from a first person's view. The reason is that the approach is able to learn similarities/differences between images that look different/similar. This is achieved using metric learning based on triple loss \cite{schroff2015facenet}:
\begin{equation}\label{tcn-triple-loss}
||f(x_i^a) - f(x_i^p)||_2^2 + \alpha < ||f(x_i^a) - f(x_i^n)||_2^2, \: \forall \{f(x_i^a), f(x_i^p), f(x_i^n)\} \in \mathcal{T}
\end{equation}
where $x_i^p$ and $x_i^n$ are the positive and negative samples (i.e. video frames) and $x_i^a$ is the anchor sample. A pair of positive and anchor samples show the same point in time from different viewpoints. The negative sample belongs to the same video containing the anchor sample but the time of the negative sample is different from the anchor sample. The embedding processes of $x_i^p$, $x_i^n$, and $x_i^a$ are computed using $f(x) \in \mathbb{R}^d$, which is modeled using a DNN. The parameters of $f(x)$ are learned such that the distance between positive and anchor samples is decreased and the distance between negative and anchor samples is increased. In Equation (\ref{tcn-triple-loss}), $\alpha$ is the minimum margin enforced between pairs of positive and negative examples.

DL methods rely on visual feedback to sense the object state requiring camera(s) calibration \cite{openai2018learning, nagabandi2020deep} or motion capture systems \cite{akkaya2019solving}. Moreover, visual data cannot provide contact and force feedback which are vital for dexterous manipulation. Alternatively, while tactile sensors can be used for object state estimation, this approach is not without challenges either. CNN has been used to capture tactile signature of human hand during grasping \cite{sundaram2019learning, funabashi2020stable} but it expects a rectangular input, which does not necessarily match the pattern of tactile sensors. Forcing rectangular pattern by rearranging the tactile readings violates their spatial relationships. To remedy this issue, graph convolutional networks (GCN) \cite{kipf2016semi} were proposed. GCN assumes that the number of tactile signals and their connections are fixed during robot operation. However, robotic hands can take different poses, causing time-variant spatial relationships between its tactile sensors which GCN fails to handle. The limitation of GCN is addressed by treating tactile signals as 3D position of activated sensors. A set of features at different levels \cite{qi2017pointnet++} is then computed based on these 3D points using a hierarchical perception model called the tactile graph neural network(TacGNN) \cite{yang2023tacgnn} which is based on a graph neural network (GNN) \cite{sanchez2021gentle}. TacGNN can handle dynamic number of tactile signals and their connections. Using TacGNN for object pose prediction, PPO has been used to learn in-hand manipulation.
\section{Simulation to real transfer}\label{sim2real-sec}
As manipulator robots become more sophisticated, their control becomes more complex. Learning methods can be used to achieve better control modules with fewer hand coded components. RL agents can learn from raw interactions with their environments without needing labeled training samples. However, during interaction, random actions are often performed to explore the problem search space, which is not practical on real robots. The reason is that sending random commands to robot joints causes jerky motions. This can damage the robot servo motors \cite{haarnoja2018learning, ibarz2021train, rodriguez2021deepwalk}. Moreover, running RL agents on real robots is time consuming, which also requires human supervision due to safety reasons. Instead of running the learning process on real robots from scratch, it is possible to learn control policy in high precision robotic simulators such as Gazebo \cite{gazebo-sim} and MuJoCo and fine-tune the learned policy on real robots \cite{smith2022legged}. However, there is always a gap between the dynamics of the simulator and that of the real world \cite{heess2017emergence}. Zhang et al. \cite{zhang2015towards, zhang2016modular} modified DQN by partitioning its structure into perception and control modules connected via a bottleneck layer. Due to the bottleneck layer, the perception module is forced to learn a low dimensional representation of the environment state, whereas the control module is responsible for Q-value estimation. The first training phase of the perception module is conducted in a supervised manner using simulated data. The second training phase is fine-tunning based on a limited number of real samples.

The gap between simulation and real world can also be addressed using dynamics randomization \cite{peng2018sim,akkaya2019solving}. In this method, the RL agent is exposed to several environments with dynamics models randomly sampled from a specific distribution. This way the RL agent has no choice but to learn a policy capable of coping with various environments with different dynamics. Hopefully, the learned policy is more likely to generalize to dynamics of the real world. The dynamics randomization is inspired by well-established field of domain adaptation \cite{tzeng2020adapting,gupta2017learning,wulfmeier2017mutual} and domain generalization \cite{wang2022generalizing}.

\begin{figure}[] 
\centering
\includegraphics[width=0.4\textwidth]{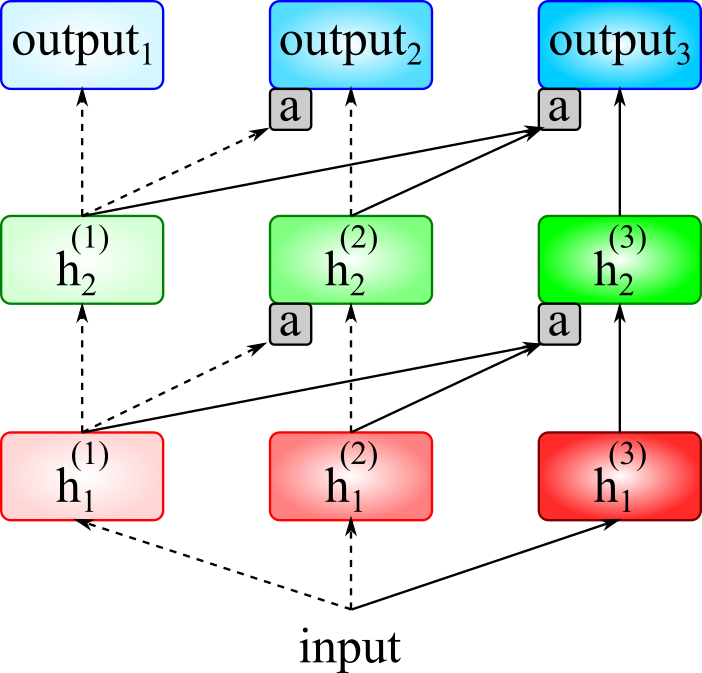}
\caption{Illustration of a three column progressive network}
\label{progress-net-fig}			% label for the whole figure
\end{figure}

In another attempt to achieve generalization, Shahid et al. \cite{9282951} trained a PPO agent to carry out grasping behaviors using a robot arm. The generalization was achieved by proposing a specific dense reward function. However, the experiments were solely done in simulation. To bridge the gap between simulation and reality of manipulation tasks, Rusu et al. \cite{rusu2017sim} used progressive networks \cite{rusu2016progressive}. These networks can learn multiple independent tasks by leveraging knowledge transfer and avoiding catastrophic forgetting \cite{kemker2018measuring, mccloskey1989catastrophic}. The difference between progressive networks and fine-tuning is that some of the previously learned knowledge is discarded during fine-tuning which has a destructive effect to some extent. Moreover, fine-tuning assumes that the tasks are partially related so only the last layer of the pretrained model is usually retrained \cite{yosinski2014transferable}. Such limiting assumption is not made by the progressive networks. Initially, the progressive network has a single column which is a neural network with L hidden layers destined to learn a specific task. Each new task is represented by a new column added to the architecture. The activation function for the ith layer of the jth column of the network is $h_i^{(j)} \in \mathbb{R}^{n_i}$ where $n_i$ is the number of units at layer $i \le L$. The parameters set of the jth column is denoted as $\theta^{(j)}$. After learning the first task, $\theta^{(1)}$ is frozen, while the second column is added to the model with parameters set $\theta^{(2)}$ for learning the second task. The output of $h_i^{(j)}$ depends on activations of columns $\{1, ..., j\}$ from previous layer $i-1$. Instead of modifying the parameters of a previously trained model, progressive networks express each learned task as a column in their architecture. Owing to the lateral connections, progressive networks are free to use or ignore knowledge of previous tasks to learn a new task. The lateral connections are depicted by "a" in Fig. \ref{progress-net-fig} and they are expressed as an MLP with one hidden layer. The activations from the previous layer are multiplied by a trainable scalar to adjust the amount each column $\{1, ..., j-1\}$ from the previous layer contributes to $h_i^{(j)}$. The fact that the progressive networks keep all of the previously learned models and establish useful connections between them to form a feature hierarchy makes them suitable for lifelong learning \cite{thrun1995learning}.

% Generalizing from Simulation 2017: https://openai.com/blog/generalizing-from-simulation/

Inverse dynamics model has also been used for simulation to real transfer \cite{christiano2016transfer}. The policy $\pi_{source}$ for robot control in the source domain (simulation) is assumed to be known. Additionally, it is assumed that the source and target (real world) domains have the same number of actuated DoFs. An inverse dynamics model $\phi(o_{-k:}, \hat{o}_{next})$ is learned to predict action $a_{target}$ to be executed in the target domain. The prediction is conducted such that executing $a_{target}$  in the target domain given the last $k$ observations $o_{-k:}$ leads to observation $\hat{o}_{next}$. Cascading source policy $\pi_{source}$, forward dynamics model $T_{source}$ and the learned inverse dynamics model $\phi$ yields an appropriate policy for the target domain.

% TODO: talk about references below:

\section{Manipulators in action}\label{manip-in-action-sec}
Over the years, manipulators have undergone constant improvements making their deployment in various application fields possible. This section is devoted to some notable applications.
\subsection{Industrial applications}
Ever since the industrial revolution, establishing an efficient workforce has always been a concern and human labor is only part of the solution. In manufacturing pipelines, automating repeated tasks such as part inspection and assembly is vital to have cost-effective production. DL has proved useful in the aforementioned automation tasks based on vision sensors. As an example, defective forged aluminum rims were detected using YOLO \cite{redmon2016you,redmon2017yolo9000,redmon2018yolov3}. Following the footsteps of Dewi et al. \cite{dewi2021yolo}, the training dataset size was increased using synthetic samples generated by GAN and DCGAN \cite{radford2015unsupervised, karras2019style}. In some cases, manipulators are equipped with laser scanners for precise measurement and inspection in production lines \cite{kuts2016robot}. The inspection process can be made more robust and adaptive using RL \cite{brito2020machine}.

Robots are also used for interacting with objects such as pick-and-place \cite{tsai2014hybrid} or fine manipulation \cite{cutkosky2012robotic}. Mitsubishi Movemaster RV-M1 was a manipulator accompanied with built-in webcam to carry out pick-and-place \cite{djajadi2010model}. The experiments showed that performance of the system was higher by 20\% if objects were colored \cite{ali2018vision}. The pick-and-place task is more challenging for micro-objects \cite{zhang2009autonomous}. In high precision manufacturing assembly, fine manipulation is a necessity which can be addressed by force-based control strategies. A typical example of such strategies have been used to carry out peg-in-hole task \cite{van2018comparative}. Manipulator can also be used for automated packaging of products such as shoes \cite{gracia2017robotic} and foods \cite{chua2003robotic}.

Traditionally, manipulators have been fixed in specific stages of industrial production pipelines. However, mobile manipulators can change their operational environment on demand increasing productivity and decreasing the number of required manipulators. CARLoS \cite{carlos_robo} developed by Robotnik is a mobile welding robot operating in shipyards.

In industrial environments, robots have to operate alongside human workers. Therefore, robots must be designed in a way that human safety is guaranteed. Different sensors such as LiDAR and cameras \cite{zhou20193d} can be used to sense the surrounding of a robot, in order to avoid collision with obstacles including humans.

In a manufacturing pipeline, certain fixed processes are designed which are repeated over and over again. However, having a flexible pipeline to adapt to predicted/unpredicted changes is highly desirable. This requirement is addressed by flexible manufacturing systems (FMS) \cite{browne1984classification}. Common components of a typical FMS are computer numerical controlled machines (CNCs), robots, sensors, etc. To setup an efficient FMS, assessment and selection of appropriate set of robots for collaboration with each other is crucial. Having a framework \cite{culleton2017framework} to carry out such assessment can facilitate the FMS setup.
\subsection{Agricultural applications}
Gardeners and farmers usually prefer to harvest their crops using human labor. However, lack of workforce specially in situations such as COVID-19 pandemic has forced farmers and gardeners to rely on robots for corps harvesting. The primary challenge is not to damage crops e.g. fruits \cite{wang2022polynomial, davidson2020robotic, zhang2023design, au2022monash} during the harvesting process. To address this challenge, deformable objects manipulation with autonomous grasp force adjustment is needed. Hand-coding a control system with the aforementioned capabilities is usually very challenging. Learning a robust controller using MFRL \cite{salhotra2022learning, almaghoutvision} is a viable option. Moreover, automated crop harvesting requires mobile manipulators. A common solution is to mount the manipulators on mobile robots equipped with autonomous navigation technologies \cite{nahavandi2022comprehensive,nahavandi2022autonomous}.

\begin{figure}[] 
\centering
\includegraphics[width=1\textwidth]{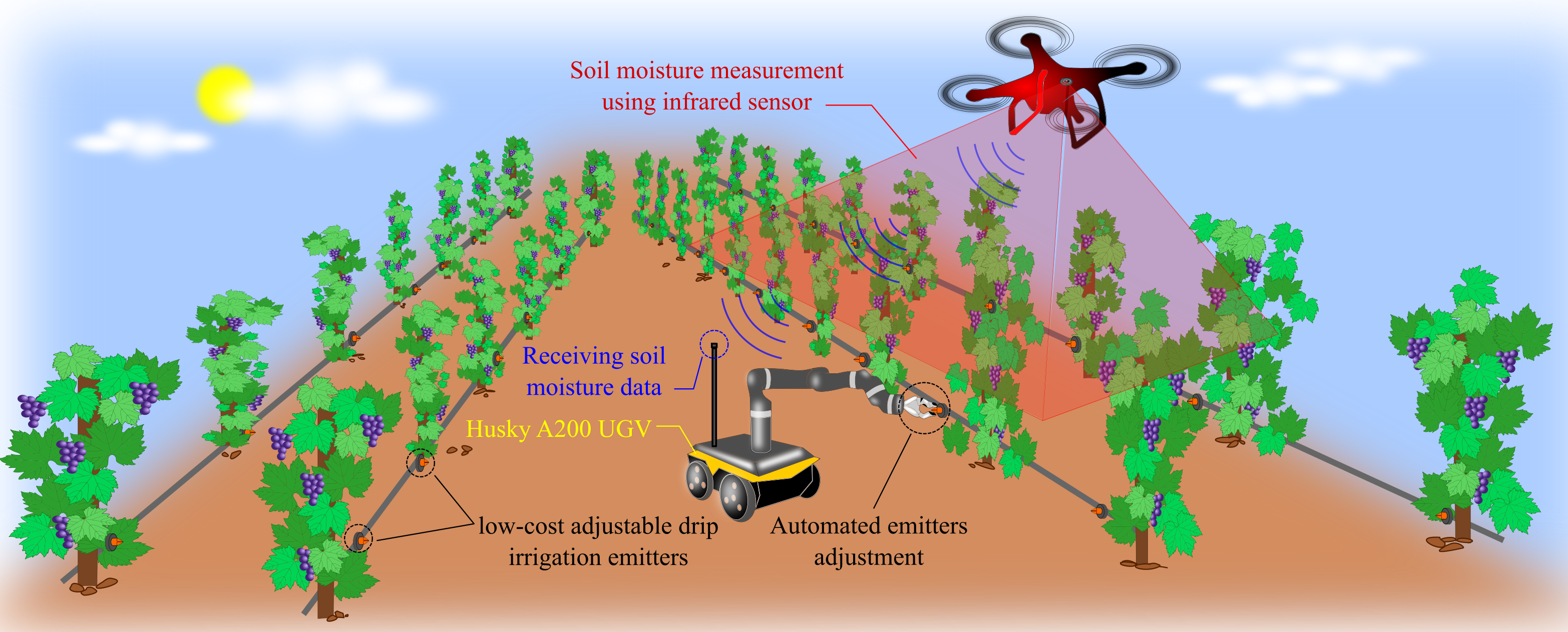}
\caption{RAPID project illustration: drone collects soil moisture data based on which Husky A200 UGV adjusts drip irrigation emitters autonomously.}
\label{rapid-fig}			% label for the whole figure
\end{figure}

Water conservation is another important use case of mobile manipulators. According to the Proceedings of the National Academy of Sciences of the USA (PNAS), about 85\% of water consumption in the world is due to agricultural activities \cite{water_usage}. That is why precision irrigation has attracted researchers' attention in recent years. UC Merced, UC Berkeley and UC Davis campuses of University of California have worked on Robot Assisted Precision Irrigation Delivery (RAPID) project to reduce water consumption in drip irrigation systems. To this end, as shown in Fig. \ref{rapid-fig}, a drone equipped with an infrared sensor is used to inspect the soil moisture condition of agricultural fields \cite{tseng2018towards}. Based on the gathered data, the drip irrigation emitters are adjusted for each plant. The adjustment can be carried out by a human workforce using a hand-held device called DATE \cite{gealy2016date} or by a manipulator arm equipped with two cameras for drip irrigation emitter detection \cite{berenstein2018robustly}. As shown in Fig. \ref{rapid-fig}, the manipulator arm is mounted on the Husky A200 unmanned ground vehicle (UGV) operating in ROS. This UGV moves throughout the field autonomously and adjusts the drip irrigation emitters for each plant.
\subsection{Military applications}
In the history of armed conflicts, many lives have been lost, but autonomous robots can make a change. It is time for UGVs to replace military personnel to carry out various hazardous tasks. Prowler and Guardium UGVs \cite{xin2013latest} can perform patrol routines. Other UGVs such as PackBot525 \cite{packbot525_robot} and Centaur \cite{centaur_robot} are capable of bomb disposal, surveillance, and reconnaissance. These two UGVs are equipped with robotic arms for effective interaction with their task environments. This way unexploded ordnance, mines, and chemical bomb can be neutralized without endangering human operators. A schematic of Packbot525 is shown in Fig. \ref{packbot525-fig}.
\subsection{Medical applications}
\begin{figure}[] % use figure* instead of figure if you want to span subfigures across two columns
\centering
\subfloat[]{\includegraphics[width=0.45\textwidth]{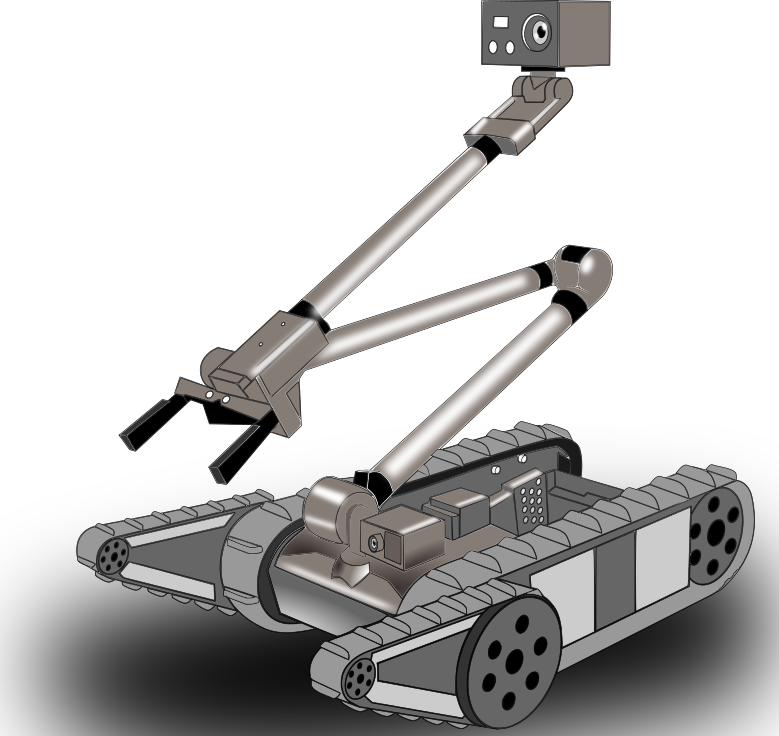}%
\label{packbot525-fig}}
\hspace*{0.9em}
\subfloat[]{\includegraphics[width=0.4\textwidth]{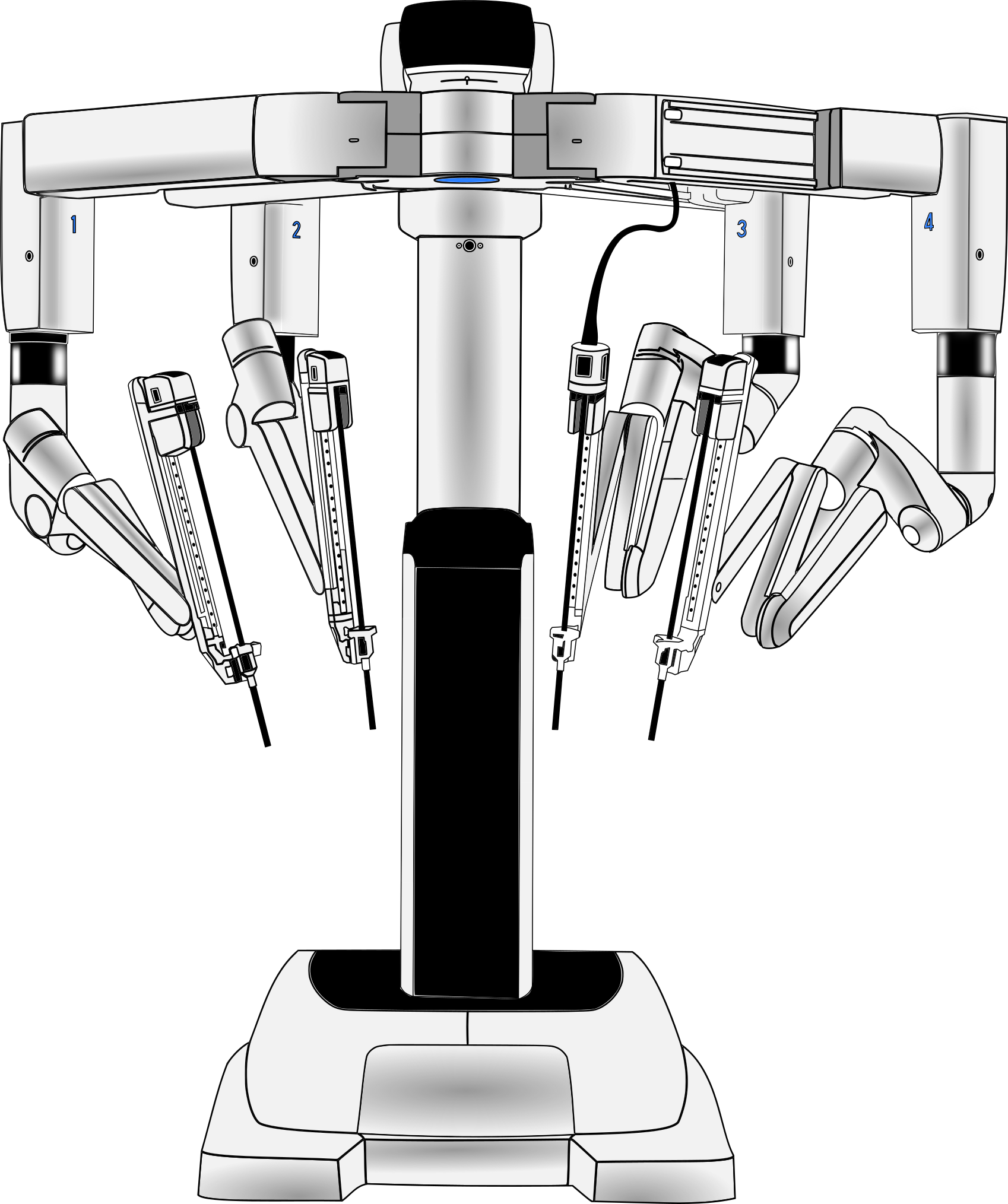}%
\label{DaVinci-fig}}
\caption{(a) PackBot525 UGV is equipped with a camera-mounted robotic arm for precise object manipulation, (b) Da Vinci surgical robot contains multiple arms for minimally invasive surgeries.}
\label{}			% label for the whole figure
\end{figure}

Surgeries are usually invasive procedures which cause side effects and discomfort for patients. Therefore, it is desirable to minimize the degree of invasion during surgeries. Da Vinci surgical robot \cite{freschi2013technical} has made minimally invasive operations possible. This robot features multiple robotic arms that can be controlled by a surgeon via its console. This robot is also accompanied by a third module, the vision cart which performs image processing, delivers live feed of the procedure to medical staff in the operation room and provides communication across Da Vinci system components. The schematic of Da Vinci robot is shown in Fig. \ref{DaVinci-fig}.

Children suffering from Autism disorder have trouble making social contacts, which leads to differed development of their social skills. Research studies show that Autism children exhibit highly organized behavior and prefer to interact with robots rather than humans. The reason is that robots are really good at repeating organized behaviors over and over without getting tired, complaining, or judging the Autistic children. The robots designed for Autism treatment have proved to be useful \cite{islam2023robot,Saleh2020RobotAF}. Examples of these robots are Nao humanoid robot \cite{mishra2023social}, Zeno R-50 \cite{salvador2015emotion}, QTrobot \cite{qtrobot}, and Kaspar \cite{robins2018kaspar}.
\subsection{Search and rescue}
Robots can play a vital role in the aftermath of natural disasters such as earthquake, tsunami, volcanic activity, etc. Rescue robots can inspect hazardous regions such as damaged nuclear sites, flooded power plants, and buildings on fire. Extensive research has been conducted on rescue robots \cite{liu2007current, liu2013robotic, 8866137, hong2022slam}. Despite advancements made in the field of autonomous rescue robots, the research still continues. To boost the progress in this field, several competitions such as the DARPA Robotics challenge \cite{krotkov2018darpa, atkeson2018happened} (2012-2015) and RoboCup \cite{kitano1999robocup, akin2013robocup} (an annual competition) have been held. In the DARPA Robotics challenge, tasks such as vehicle driving, door opening and going through, tool manipulation and some other complex tasks had to be completed by robots autonomously. All these tasks demand mobile robots with robust dexterous manipulation capabilities. The DRC-HUBO robot \cite{lim2017robot} developed by team KAIST managed to win the competition in 2015.
\subsection{Space}
An extravehicular activity (EVA) is any type of activity that is done in the vacuum of space by an astronaut. EVAs are always challenging and risky but these activities can be carried out by robotic manipulators to reduce the risk and time of space missions. Robonaut 2 (R2) \cite{5979830, tzvetkova2014robonaut} is a dexterous humanoid robot developed by NASA and General Motors \cite{badger2016ros}. Initially, R2 only had the upper-body parts but was equipped with a mobility platform in order to maneuver around the international space station (ISS). R2 can be used for automatic maintenance and cleaning tasks such as handrail cleaning, vacuuming filter, inventory management, and data collection inside the ISS. Robot operating system (ROS) \cite{quigley2009ros} tools and capabilities have been used extensively to develop R2 controller system.

{Manipulators may also be used for removing debris due to 5000 rockets launched from Earth into the orbit successfully. Typical debris include defunct spacecrafts, spent rocket upper stages \cite{bombardelli2013multiple, aslanov2020spent} and other fragments. Debris wandering in space is a serious threat to functional satellites in orbit.

In addition to the applications mentioned above for space manipulator, rovers equipped with manipulator arms can act as astronaut's assistants. These rovers can explore hazardous planets and gather data during their mission or repair a malfunctioning on-board lander \cite{zaman2022phoenix}. Moreover, the required technology for development of these rovers can also fulfill the requirements of industry 4.0.

Regardless of tasks assigned to space manipulators, their structural characteristics may vary slowly over time. To cope with this issue, manipulator controllers must be adaptable to uncertainties due to the aforementioned structural changes. To this end, one may use a weighted mixture of multiple Kalman filters to setup a robust controller \cite{fekri2006issues} or utilize RL to learn an adaptable controller \cite{pradhan2012real}.

% some more ref:
% * How scientists reviewed the process and development of space intelligent robot technology? 2022
% 		** https://www.eurekalert.org/news-releases/948130
% * Tutorial Review on Space Manipulators for Space Debris Mitigation
% 		** https://www.mdpi.com/2218-6581/8/2/34
\section{Future works}\label{future-work-sec}
As manipulation tasks get more complex, the need for robust and efficient robot controllers based on ML increases. As discussed before (section \ref{enf-robot-user-safe-sec}), the number one requirement for using learning in real world problems is robot and user safety enforcement. Safe HRL is already under development for robot navigation \cite{xiong2022hisarl, zhu2022hierarchical, gangopadhyay2021hierarchical} and learning arcade games \cite{jain2021safe}. However, it still needs further research in robotic manipulation tasks.

Another future work is related to the rigidity of objects. In dexterous manipulation, the objects of interest are usually assumed to be rigid. However, many real-world applications require interaction with non-rigid objects \cite{huo2022dual}. Devising HRL methods with the ability to handle non-rigid objects is an important research direction for the future.

While visual perception is popular for dexterous manipulation, tactile-based sensing provides better feedback for tasks such as finger gating, grasping, and in-hand manipulation. The reason is that tactile sensors integrated in robot hands provide feedback about fingers breaking/establishing contacts with objects. Such observations are vital to detect key transitions \cite{johansson2009coding} in the problem state space but these observations are not provided by visual sensors. The limitation of tactile-based sensing is blind spot in the robot hand in which no tactile sensors are present. Getting object status in blind areas is an important future work. Relying on time series data, it may be possible to approximate the object status based on previously observed data \cite{yang2023tacgnn}.

Human-Robot Collaboration (HRC) has attracted attention in recent years. This stems from the fact that combination of humans' creativity and accurate robots has the potential for efficient production and personalization capabilities in industrial applications. To establish HRC efficiently, robots must be able to perceive and analyze information obtained from their workspace in real-time. This holistic scene understanding involves object perception, environment parsing, human recognition, and visual reasoning which leads to better collaboration between humans and robots. As future work for HRC, better cognition skills for robots must be developed. The ability to take appropriate actions for manipulation of objects never seen before is crucial. Examples are picking-and-placing unknown objects \cite{zeng2022robotic} as well as deducing whether a liquid can be poured into an unknown container \cite{wu2020can}. Visual perception plays major role for perception of the robots' workspace. Typical steps of visual parsing of the environment is its recognition using vision algorithms and representing scene elements based on mapping techniques (e.g. 2D, 3D representation). However, these techniques does not take into account hierarchical and hybrid representation of the robot workspace which is needed for better modeling of flexible manufacturing lines. Therefore, hierarchical modeling of the workspace is worthy of further research \cite{fan2022vision}.
\section{Conclusion}\label{conc-sec}
We are living in an exceptional era of rapid technological advancements in which robotic manipulators play a major role in achieving efficient and reliable automation. While traditional hand-coded controllers have been around for a long time, the ever-increasing complexity of robots demand better controllers that are easier to work with. ML can be used to abstract away the complexity of developing hand-coded controllers to some extent. This is achieved by exploiting domain knowledge in the form of expert's demonstration and/or using RL for learning controllers via interaction with the robot's workspace. However, developing reliable controllers based on ML still needs further research. Robustness against adversarial attacks for DL-based controllers, enforcing human safety around the robots, and achieving cost-efficient training are just some of the factors that can pursued in future works.

In industrial domain, the input to DL models are usually sensor readings of various types. As long as sensors are functional, the DL models do the objective they are designed to do. Now suppose one or more of the sensors malfunction due to wear and tear. From the DL models view point, the invalid readings of faulty sensors are like OOD inputs which may corrupt the outputs of the DL models in unpredictable ways. This behavior can be disastrous in safety critical applications. Therefore, it is vital to develop DL models that can detect OOD inputs.

In profitable industries, minimizing the cost and the production lines down time due to maintenance is very important which is achievable via predictive maintenance. AI methods can be used to implement predictive maintenance effectively \cite{rojek2023artificial}. More generally, McKinsey Global Institute has forecasted in its 2018 report \cite{bughin2018notes} that AI has the potential to contribute about 13 trillion dollars to the economy worldwide by 2030. In addition to being profitable, industries must be sustainable i.e. they must minimize their impact on our environment. AI can be utilized to move toward green technologies. One of the critical sectors that AI and robotics can be of great help is optimal energy consumption. Practical examples are optimal power consumption of Google's data centers using ML \cite{lazic2018data} and green manufacturing by replacing humans with robots. The latter case leads to lower energy consumption \cite{zhang2022robot} and significant reduction in carbon emission \cite{li2022carbon}. Additionally, ML can play major role in big data analysis \cite{kapp2020pattern} and realtime performance monitoring of large scale systems \cite{may2021foresighted} which is vital in the competitive market of Industry 4.0. This survey tried to sum up the state-of-the-art ML-based approaches for controlling manipulators. Moreover, real-world applications of manipulators in different sectors such as military, healthcare, agriculture, space, and industry were reviewed.



%\bibliographystyle{elsarticle-num} 
%\bibliography{refs}


\end{document}